\documentclass[11pt]{article}
\usepackage{float} 

\usepackage{amsmath}
\usepackage{cite}
\usepackage{graphicx}
\usepackage{float}
\usepackage{natbib}
\usepackage{booktabs} 
\usepackage{graphicx}
\usepackage{subcaption}
\usepackage{float}
\usepackage[utf8]{inputenc} 
\usepackage[T1]{fontenc}    
\usepackage{hyperref}       
\usepackage{url}            
\usepackage{booktabs}       
\usepackage{amsfonts}       
\usepackage{nicefrac}       
\usepackage{microtype}      
\usepackage{lipsum}
\usepackage{fancyhdr}       
\usepackage{graphicx}
\usepackage{todonotes}
\usepackage{amsmath}
\usepackage{amssymb}
\usepackage[margin=0.9in]{geometry} 
\usepackage{booktabs}  
\usepackage{caption}   
\usepackage{booktabs}
\usepackage{multirow}
\usepackage{graphicx}
\usepackage{subcaption}
\usepackage{float}
\usepackage{amsmath, amssymb}
\usepackage{bm}
\usepackage{amsmath}
\usepackage{graphicx}
\usepackage{enumitem}
\usepackage{hyperref}
\usepackage{adjustbox}        
\usepackage{natbib} 
\usepackage{xcolor}

\definecolor{reviewblue}{RGB}{0,0,0} 
\newenvironment{reviewtext}{\color{reviewblue}}{}
\graphicspath{{media/}}     

\title{Informed Forecasting: Leveraging Auxiliary Knowledge to Boost Large Language Models Performance on Time Series Forecasting}

\author{
  Mohammadmahdi Ghasemloo$^1$, Alireza Moradi$^2$ \\
  $^1$Texas A\&M University, College Station, TX, USA \\
  $^2$Georgia Institute of Technology, Atlanta, GA, USA \\
  \texttt{mohammad\_ghasemloo@tamu.edu, alirezamoradi@gatech.edu}
}

\begin{document}

\maketitle

\begin{abstract}
The rapid adoption of large language models (LLMs) has sparked growing interest in extending their capabilities beyond traditional natural language tasks, including applications in time series forecasting. This work explores the enhancement of time series forecasting with LLMs by incorporating time-dependent covariates. We propose a set of representative prompting strategies that span a wide range of formats while incorporating covariates, and \textcolor{reviewblue}{then implement the proposed framework to evaluate the performance across three real-world time series datasets in healthcare, service operations, and transportation.} Our experiments demonstrate that incorporating correct covariate information through a suitable prompt design can significantly improve forecast accuracy. \textcolor{reviewblue}{Furthermore, the analysis reveals that both the choice of covariate and the prompt structure are critical, as poorly aligned configurations may degrade performance. Finally, we conduct sensitivity analyzes to assess the effect of covariate integration in censored settings and quantify uncertainty, which confirms that our method achieves statistically significant improvements over existing approaches. These findings underscore the potential of covariate integration in prompt design to bridge the gap between general-purpose LLMs and forecasting tasks.}

\end{abstract}

\vspace{1em}

\section{Introduction}\label{sec:intro}

The emergence of Large Language Model (LLM) services has led to a rapid increase in their global usage. Platforms such as ChatGPT, for example, report more than 300 million weekly users, which underscores the widespread integration of these tools into daily life and professional workflows \citep{chatgpt_growth_2024}. \textcolor{reviewblue}{This rapid adoption has, in turn, catalyzed efforts to repurpose LLMs beyond core NLP settings.}

LLMs originally designed for the understanding and generation of natural languages have quickly expanded to a wide range of complex domains. Recent research has demonstrated their promising capabilities in areas such as robotics control, optimization, simulation-based reasoning, task-oriented dialog systems, and time series forecasting~\citep{jin2023time, akhavan2024generative, vemprala2024chatgpt, tang2025time, yu2023temporal, meem2024outofschema, karimian2024explainable}. \textcolor{reviewblue}{Building on this trend, we focus specifically on the role of LLMs in time series forecasting.}

Time series forecasting plays a vital role in decision-making across a range of application domains. In healthcare, accurate forecasts of disease incidence can inform resource allocation, staffing, and public health interventions. In service operations, demand forecasting enables better capacity planning and scheduling. In finance, reliable forecasts support investment strategies and risk management. \textcolor{reviewblue}{Given these stakes, it is natural to ask when and how general-purpose LLMs can contribute to time series forecasting in a useful and low-friction manner.}

Despite the existence of specialized time series forecasting models, there are compelling reasons to explore the utilization of commonly used LLMs in this domain. First, LLMs offer a highly accessible and low-barrier alternative for non-expert users, enabling a broader community to perform forecasting tasks without modeling expertise. \textcolor{reviewblue}{Second, LLM-based forecasting is particularly well-suited for use as an initial decision-support tool in operational settings such as healthcare (e.g., predicting patient inflow), call centers (e.g., estimating daily arrival volumes), and transportation (e.g., forecasting airline passenger demand), where fast, interpretable, and adaptive forecasts can assist in planning and resource allocation.} Third, due to their fast inference and natural language interface, LLMs can be used to generate preliminary forecasts that serve as input models for high-fidelity decision tools, including simulation-based systems. \textcolor{reviewblue}{These practical advantages motivate a closer look at design choices for LLM-based time series forecasting.}

In classical time series forecasting, covariates such as calendar attributes, weather conditions, or economic indicators play a crucial role in improving predictive accuracy. However, unlike traditional forecasting methods, most LLM-based approaches rely on raw inputs and overlook other information available in the dataset \citep{xue2023promptcast}. Recent work has started to explore this direction. \cite{xue2023promptcast} augmented LLM inputs with simple temporal covariates such as data entry date. \cite{tang2025time} emphasized the effect of external knowledge by embedding contextual attributes derived from the dataset itself. Similarly, \cite{xiao2025retrieval} proposed a retrieval-augmented framework in which relevant financial indicators are selected and included in the input prompt.

\textcolor{reviewblue}{Together, these efforts establish the value of utilizing side information but leave three questions open.} 
First, there is limited guidance on how users should formulate effective prompts to incorporate covariates into the model. More specifically, what constitutes a good prompt for presenting covariate information in a way that enhances predictive performance without overwhelming or misleading the model?
Second, a systematic approach is needed to evaluate the effect of covariate integration on the accuracy of forecasting. In other words, how can we determine which variables contribute meaningfully to predictive accuracy and how can this selection process be made robust across different datasets and forecast horizons?
Third, considerations such as data leakage and computational cost are critical along with forecast accuracy. Therefore, how can users effectively mitigate the risk of data leakage while also accounting for cost?

To address this gap, this paper introduces a pipeline for informed LLM-based time series forecasting that examines the role of time-dependent covariates available within the dataset. \textcolor{reviewblue}{To ensure practical generality, we focus on covariates that are not tied to a specific dataset, chiefly calendar-derived attributes that can be defined for virtually any series and employ a validation-based framework to evaluate each prompt and covariate in terms of forecasting performance.}

\begin{figure}[!t]
    \centering
    \includegraphics[width=0.8\linewidth]{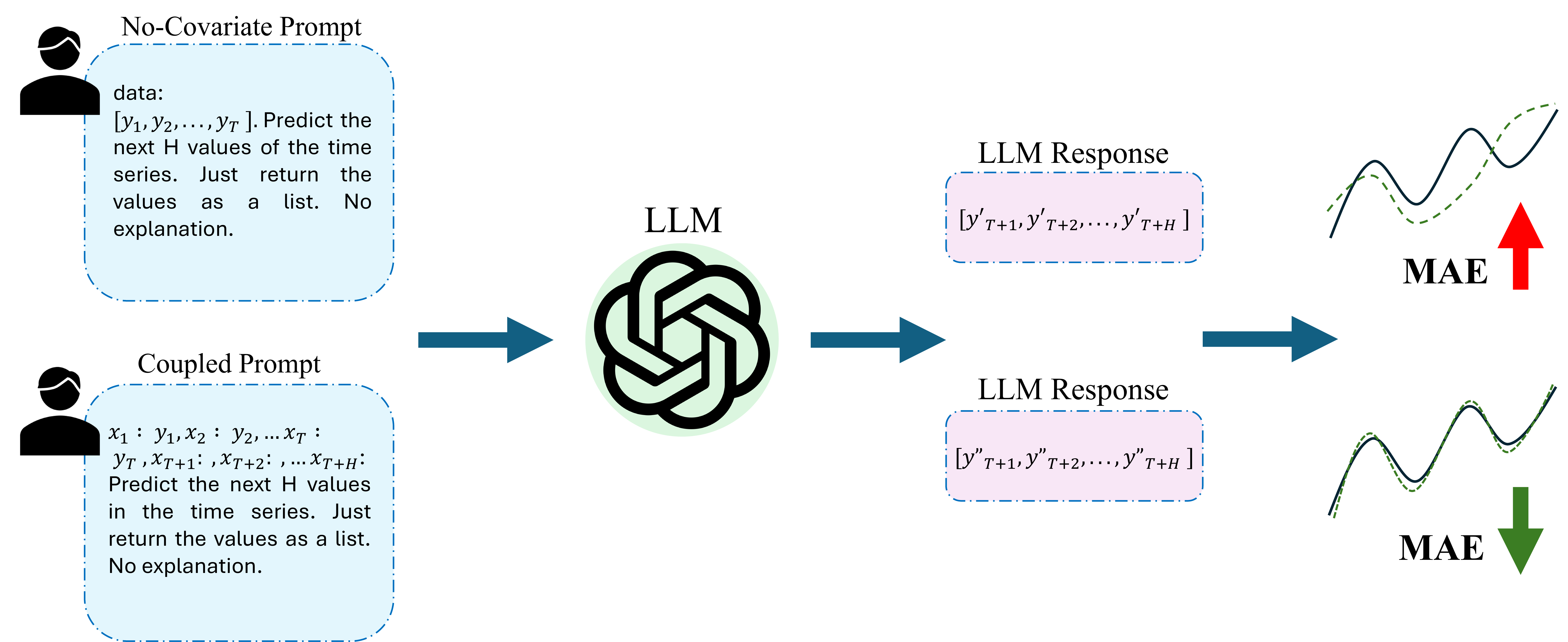}
    \caption{Overview of the process of integrating covariates into the task of forecasting with LLMs. Time-dependent covariate is integrated with raw time series data to generate forecasts through natural language interaction.}
    \label{fig:flow}
\end{figure}

We provide a curated set of representative prompt formats and implement it on three real-world datasets. These experiments demonstrate how incorporating time-dependent covariates into prompt design can significantly improve forecast performance.  
We do not claim to have discovered the optimal prompt for all scenarios, nor that our approach outperforms traditional time series forecasting methods in general. Rather, we argue that when commonly used LLMs are used for forecasting, their performance can be significantly enhanced by incorporating covariates into the prompt design. \textcolor{reviewblue}{Our objective, therefore, is to surface effective, reusable patterns for covariate integration rather than to replace specialized time series forecasting models.}  
Figure~\ref{fig:flow} illustrates the end-to-end forecasting workflow.

The remainder of this paper is organized as follows.
Section~\ref{sec:ts_review} reviews the foundational work in classical and modern time series forecasting. Section~\ref{sec:review} surveys recent developments in LLMs for forecasting tasks. Section~\ref{sec:method} introduces our proposed pipeline for leveraging LLMs in time series forecasting. Sections~\ref{sec:exp} and \ref{sec:res_n} evaluate the effect of incorporating covariates in three real-world data sets, and finally, Section~\ref{sec:conclusion} summarizes the key findings and provides directions for future research.

\section{Time Series Forecasting}\label{sec:ts_review}
Time series forecasting has long been one of the most attractive and widely applicable areas of research for data-driven fields. Consequently, significant efforts have been made by the research community to advance this domain \cite{kolambe2024forecasting}.
The potential applications of time series forecasting are vast and diverse, encompassing domains such as finance \citep{sezer2020financial,dingli2017financial}, healthcare \citep{kaushik2020ai}, power systems \citep{koivisto2019using}, and supply chain management \citep{aviv2003time}, among others. This wide range of use cases underscores the importance of time series forecasting in real-world operations and decision-making.
As a result, numerous methodologies have been developed for time series prediction, ranging from classical statistical approaches like ARIMA \citep{ariyo2014stock,fattah2018forecasting,chen2008forecasting}. For example, \cite{ariyo2014stock} applied ARIMA to forecast the Nigerian stock market, while \cite{fattah2018forecasting} used it to forecast electricity demand with seasonal adjustments.
Linear prediction models have also been extensively employed in forecasting tasks \citep{yildiz2017review,bianco2009electricity}. \cite{yildiz2017review} presented a comprehensive review of linear and hybrid models in the prediction of energy systems, and \cite{bianco2009electricity} applied multiple linear regression to predict electricity demand in Italy.

With the advancement of computational capabilities, machine learning techniques have gained attraction in time series forecasting \citep{obata2021random,banskota2014forest}. \cite{obata2021random} demonstrated the effectiveness of Random Forests in predicting electricity load with high-dimensional input, while \cite{banskota2014forest} applied machine learning models to predict forest disturbances using remote sensing data. More recently, deep learning frameworks have emerged as powerful tools for modeling complex temporal patterns in time series data \citep{sezer2020financial,zeroual2020deep}. \cite{sezer2020financial} reviewed deep learning approaches such as CNNs and LSTMs in financial time series forecasting, while \cite{zeroual2020deep} performed a comparative analysis of deep learning models in various forecasting tasks and highlighted their superior performance over traditional methods. In addition, \cite{karimian2024explainable} proposed an explainable deep learning architecture for multivariate time series forecasting that results in interpretability with high predictive accuracy.

\section{LLMs as Time Series Forecasting Models}\label{sec:review}

LLMs have found significant applications in scientific research, expanding their role beyond their initial design \citep{liang2024mapping}. Although the primary function of LLMs is to process, understand, and generate human-like text by identifying patterns and relationships in vast datasets \citep{ibm_large_language_models}, recent research has explored their potential for applications that go beyond their original purpose.
Several recent works have initiated this exploration. For instance, \cite{jin2023time} introduced Time-LLM, a framework that reprograms pre-trained LLMs for time series forecasting without retraining by transforming numerical data into textual prompts. Similarly, \cite{gruver2023large} demonstrated that models such as GPT-3 and LLaMA-2 can forecast time series in a zero-shot setting without prior domain-specific training. \cite{chang2025llm4ts} proposed LLM4TS, which aligns the LLM with the time series using a two-stage fine-tuning strategy. More recent work incorporates multimodal or external knowledge. \cite{xiao2025retrieval} introduced a retrieval-augmented LLM for financial forecasting that combines time series and financial indicators, while \cite{jia2024gpt4mts} integrates numerical and textual data for multimodal prediction. Similarly, \cite{yu2023temporal} explore explainable LLM-based forecasting by combining time series, metadata, and news. Despite these advances, most methods include covariates in a fixed way without systematically exploring which combinations are most effective. Existing approaches, such as PromptCast proposed by~\cite{xue2023promptcast}, which adapts prompt-based learning to structured time series tasks, and the method introduced by~\cite{tang2025time}, which investigates LLM performance across different data patterns and proposes prompt engineering strategies to improve generalization, can both be accommodated within our framework.  
\textcolor{reviewblue}{Recent research has extended LLM-based forecasting in two notable directions. 
The first is patch-based prompting, where a time series is divided into patches and presented to a frozen LLM with structured instructions  (for example, PatchInstruct; \citep{bumb2025forecasting}).  The second is integer or discretization approaches, which transform continuous values into integer or symbolic tokens, often combined with cross-modal alignment or light fine-tuning  (e.g. IDDLLM; \citealp{wang2025novel}).  Both lines of work primarily focus on reformatting numerical inputs for LLM consumption.  These methods primarily focus on alternative representations of numerical series, while covariates are handled in a more implicit way (e.g., through patches, discretization, or textual cues). Our study complements these advances by examining explicit and systematic covariate integration in LLM-based forecasting.}


Table~\ref{table:llm_forecasting_summary} provides a summary of the use of LLMs for time series forecasting. As the table shows, our study is unique in designing a set of custom prompts to incorporate covariates such as time embeddings into the LLM input, while also conducting a systematic evaluation of different covariate combinations to identify the most effective configuration to improve predictive accuracy.

\begin{table}[h!]
\captionsetup{justification=centering}
\caption{Summary of Recent Studies on Time Series Forecasting Using LLMs}
\centering
\renewcommand{\arraystretch}{1.9}
\fontsize{10}{12}\selectfont
\begin{adjustbox}{width=\textwidth}
\begin{tabular}{@{}p{2.6cm} p{2.6cm} p{5.6cm} p{2.8cm} c c c c@{}}
\toprule
\textbf{Paper} & \textbf{LLM(s) Used} & \textbf{Contribution} & \textbf{Dataset(s)} & \textbf{External Knowledge} & \textbf{Covariate Integration} & \textbf{Prompt Selection} & \textbf{Covariate Selection} \\
\midrule

\cite{xue2023promptcast} & GPT-2, BART, T5 &
Proposed PromptCast, a prompt-based forecasting paradigm that leverages LLMs through natural language inputs to perform zero-shot and few-shot time series prediction. &
PISA (City Temp., Electricity, Visitor Flow) & Yes & Yes & No & No \\

\citep{gruver2023large} & GPT-3, LLaMA-2 &
Introduced LLMTime: reframed time series forecasting as a next-token prediction task, demonstrating LLMs’ capabilities in zero-shot settings. &
29 benchmark datasets & No & No & No & No \\

\citep{jin2023time} & GPT-2, BERT, LLaMA-7B &
Proposed Time-LLM: reprograms LLMs via prompt templates, achieving effective forecasting without altering model weights. &
ETT, M4 & No & No & No & No \\

\citep{chang2025llm4ts} & GPT-2 &
Presented LLM4TS, a two-stage fine-tuning framework to align pre-trained LLMs with time series forecasting tasks with minimal data. &
Seven real-world datasets & No & No & No & No \\

\citep{tang2025time} & GPT-3.5-turbo, GPT-4-turbo, Gemini-Pro, LLaMA-2-13B &
Analyzed LLMs’ strengths in trend/seasonal data and proposed methods to enhance forecasting via prompt engineering and contextual augmentation. &
Darts (Air Passengers, ETT, etc.) & Yes & No & Yes & No \\

\citep{xiao2025retrieval} & StockLLM (fine-tuned LLaMA3.2-1B) &
Developed a RAG framework for financial forecasting; includes FinSeer, an LLM-guided retriever to extract relevant historical sequences. &
Stock prices + 20 financial indicators & Yes & No & No & No \\

\citep{yu2023temporal} & GPT-4, Open LLaMA &
Demonstrated that LLMs can generate explainable forecasts by integrating multi-modal financial inputs, including time series, metadata, and news. &
NASDAQ-100 (prices, metadata, financial news) & Yes & No & No & No \\

\citep{liu2024timecma} & GPT-2  &
Proposed TimeCMA, an LLM-empowered framework for multivariate time series forecasting that aligns time series and prompt embeddings via cross-modality alignment. &
8 real-world multivariate time series datasets & No & No & No & No \\

\citep{pan2024s} & GPT-2 &
Proposed S$^2$IP-LLM, a framework that aligns time series embeddings with LLM semantic space via cross-modality alignment, using semantic anchors as prompts. &
Multiple benchmark datasets & No & No & No & No \\

\citep{jia2024gpt4mts} & GPT-2 &
Proposed GPT4MTS, a prompt-based LLM framework integrating numerical and textual data for multimodal time series forecasting. &
GDELT-based multimodal time series dataset & Yes & No & No & No \\ 

\textcolor{reviewblue}{\citep{bumb2025forecasting}} & \textcolor{reviewblue}{LLaMA-family (frozen)} &
\textcolor{reviewblue}{Patch-based prompting with structured instructions; time series are tokenized into patches, enabling forecasting without heavy fine-tuning.} &
\textcolor{reviewblue}{Common time series forecasting benchmarks} & \textcolor{reviewblue}{Optional} & \textcolor{reviewblue}{Implicit (via patches)} & \textcolor{reviewblue}{Manual/Validation} & \textcolor{reviewblue}{No} \\

\textcolor{reviewblue}{\citep{wang2025novel}} & \textcolor{reviewblue}{LLM (fine-tuned)} &
\textcolor{reviewblue}{Integer/decimal discretization of numeric series with cross-modal alignment; values mapped to integer tokens for LLM forecasting.} &
\textcolor{reviewblue}{Multiple real datasets} & \textcolor{reviewblue}{Optional} & \textcolor{reviewblue}{Implicit (via discretization)} & \textcolor{reviewblue}{Learned} & \textcolor{reviewblue}{No} \\

\textbf{This Study} & \textbf{GPT-4o-mini} &
\textbf{Designed a novel prompting strategy that explicitly encodes external covariates (e.g., time embeddings) into LLM inputs, and systematically investigates which combinations yield the best forecasting performance. Evaluates across multiple forecast horizons, datasets, and prompt structures.} &
\textbf{WHO FluNet, Oakland Call Center} & \textbf{Yes} & \textbf{Explicit (horizon-aligned)} & \textbf{Validation} & \textbf{Validation} \\

\bottomrule
\end{tabular}
\end{adjustbox}
\label{table:llm_forecasting_summary}
\end{table}

\section{Proposed Framework}\label{sec:method}
In time series forecasting context, three quantities define the context: The observed data, which consists of historical target values; the forecast horizon, which specifies how many future time steps are to be predicted; and the covariates, which are known exogenous variables aligned with the time series that may influence future outcomes and are available both for the observed data and forecast horizon.
Our objective is to predict future values \( y_{T+1}, \dots, y_{T+H} \) given a history of past observations \( y_1, \dots, y_T \) and the covariates \( \mathbf{x}_1, \dots, \mathbf{x}_{T+H} \) where $T$ and $H$ denote the length of the observed data and the forecast horizon, respectively.
To assess model performance, we partition the dataset into a validation set and a test set, where the validation set is used to choose the best prompt-covariate pair.
A key step is the construction of prompts that encode these three core components into natural language.

\subsection{Prompt Structures for Forecasting}
\textcolor{reviewblue}{Our objective is to design a set of prompts that are deliberately simple. Specifically, we focus on prompts that rely solely on observed time series data and avoid the use of external knowledge or example completions.
Each prompt should include an instruction or a natural language description of the forecasting task, together with encoded data that contains previous observations and covariates in a structured format. Then it should conclude with a clear forecast request that directs the model to generate predictions for the specified forecast horizon.
 }

\textcolor{reviewblue}{Although there are infinitely many ways to construct prompts for time series forecasting using LLMs, ranging from raw sequences to detailed explanatory formats, we categorize prompts based on how they organize and present target values and covariates to the model and consider three representative categories:}

\begin{itemize}
    \item \textbf{Coupled Prompt.}  
    Each observation is represented as a key-value pair, combining covariates with their corresponding target values in an ordered sequence. Forecasting is performed by adding future covariates without associated targets.  
    This format leverages the autoregressive nature of LLMs by mimicking the standard language modeling task, where the model learns to associate sequences of covariates and targets in a contiguous stream. It is simple to implement and aligns well with token-based sequence modeling, which makes it a possible fit for LLMs. 

    \item \textbf{Decoupled Prompt.}  
    Covariates and target values are separated into distinct lists. The prompt contains both target values and covariates for the observed data and asks to forecast future targets using a list of upcoming covariates.  
    This separation may allow LLM to differentiate between the input features and the forecasting target.

    \item \textbf{Contextualized Prompt.}  
    \textcolor{reviewblue}{This format extends the Decoupled structure by adding context about the covariates. The goal is to help the LLM recognize recurring patterns (e.g., seasonal or weekly cycles) by emphasizing their importance. The prompt avoids revealing dataset-specific details and instead leverages the pre-trained knowledge of the LLM through high-level information.}

\end{itemize}
\textcolor{reviewblue}{As a benchmark, we include the No-Covariate Prompt, which omits exogenous information entirely and two prompting strategies inspired by previous work:}
\begin{itemize}
    \item \textbf{No-Covariate prompt (Baseline).}  
    Only the raw sequence of target values without any accompanying temporal or contextual information is included, which serves as a baseline to evaluate the effect of including covariates.
    
    \item \textbf{PromptCast~\citep{xue2023promptcast}.}  
    This format presents the input as a compressed natural language sentence that implicitly integrates covariate and target information. It avoids rigid structural alignment between covariates and target values and instead relies on the ability of the model to interpret the sentence holistically.

    \item \textbf{Knowledge-Guided prompt~\citep{tang2025time}.}  
    This format builds on the Decoupled structure by including domain-specific knowledge, such as the subject or source of the time series (e.g., 'This series represents US electricity demand').
\end{itemize}
The prompt templates corresponding to each of the prompting strategies are provided in Table~\ref{tab:prompt-examples}.

\subsection{Prompt and Covariate Selection}

\textcolor{reviewblue}{Selecting which prompt format to use and which covariate to include is a critical design decision.} While many real-world datasets contain multiple time-dependent covariates, not all contribute equally to the accuracy of the forecast.
We adopt a validation-based approach to guide both prompt selection and covariate integration in a coordinated manner, and the prompt-covariate structure that produces the best forecast accuracy in a hold-out validation set is selected. \textcolor{reviewblue}{Although this approach is computationally expensive, it can provide a reliable basis for selecting effective prompting designs.}

Specifically, we embed each covariate in the prompt structure and assess its forecast performance on the validation set. This procedure allows us to evaluate both the standalone predictive value of each covariate and its compatibility with the chosen prompt.
By jointly considering the prompt structure and covariate effectiveness, we ensure that our final design maximizes the forecasting accuracy. \textcolor{reviewblue}{The prompt--covariate pair that results in the best validation performance is then chosen.}
\textcolor{reviewblue}{ Incorporating covariates can change the tokens given to the model and may make the forecast query more similar to previous cases with the same covariate values. The model attention can then focus on matching historical segments and may act like a similarity-weighted average over past patterns that align with the upcoming covariate. Different attention heads may capture recurring structures such as weekly or yearly seasonality. In this case, informative and well-aligned covariates sharpen attention and improve accuracy, while uninformative or mismatched covariates weaken attention and reduce performance.
However, incorporating auxiliary information (e.g., covariates) is not free and may introduce certain trade-offs.}

Data leakage is a critical concern when using LLMs for forecasting. If the prompt contains domain-specific information, it may reveal information, and the model may produce overly optimistic forecasting performance that fails to generalize to unseen future data. However, added prompt complexity may also lead to failures. In general, this is an uncontrollable and unmeasurable error, and prompts such as \cite{tang2025time} could be prone to such risk.

\textcolor{reviewblue}{In the context of LLM-based forecasting, the computational overhead of different prompt formats is generally similar, but the monetary cost is largely determined by the number of tokens processed during inference. Each prompt incurs a usage fee proportional to its token length, which includes instructions, time indices, covariate values, formatting symbols, and surrounding natural language. Since the main body of the prompt is used for presenting the time series, adding covariates approximately doubles the token count relative to simple baselines, thereby nearly doubling the cost. In large-scale deployments where thousands of forecasts may be generated daily, these differences translate into substantial financial and computational overhead. Thus, covariate integration should be prioritized only when simpler and shorter prompts do not deliver sufficient accuracy.}

\begin{table}[H]
\centering
\renewcommand{\arraystretch}{1}
\setlength{\tabcolsep}{10pt}
\footnotesize
\caption{Prompt formats used in the study.}
\begin{tabular}{p{3cm} p{11cm}}
\toprule
\textbf{Prompt Type} & \textbf{Prompt Template} \\
\midrule
\textbf{No--Covariate } &
\ttfamily

\texttt{data: [$y_1$, $y_2$, $\ldots$, $y_T$]}.
\texttt{Predict the next $H$ values of the time series.}
\texttt{Just return the values as a list. No explanation.}\\

\midrule

\textbf{Coupled } &
\ttfamily
\(\mathbf{x}_1: y_1\),
\(\mathbf{x}_2: y_2\),
\ldots
\(\mathbf{x}_T: y_T\),
\(\mathbf{x}_{T+1}\): , 
\(\mathbf{x}_{T+2}\): ,
\ldots
\(\mathbf{x}_{T+H}\):

Predict the next \(H\) values in the time series. Just return the values as a list. No explanation.\\
\midrule

\textbf{Decoupled } &
\ttfamily
Data: \([y_1, y_2, y_3, \dots, y_T]\).
Covariates: \([\mathbf{x}_1, \mathbf{x}_2, \mathbf{x}_3, \dots, \mathbf{x}_T]\).
Prediction covariates: \([\mathbf{x}_{T+1}, \mathbf{x}_{T+2}, \dots, \mathbf{x}_{T+H}]\).

Predict the next \(H\) values of the time series. Just return the prediction values as a list. No explanation.\\
\midrule

\textbf{Contextualized } &
\ttfamily
Data:\([y_1, y_2, y_3, \dots, y_T]\).
Covariates:\([\mathbf{x}_1, \mathbf{x}_2, \mathbf{x}_3, \dots, \mathbf{x}_T]\).
Prediction covariates: \([\mathbf{x}_{T+1}, \mathbf{x}_{T+2}, \dots, \mathbf{x}_{T+H}]\).
The sequence represents a univariate time series with aligned covariates. These covariates exhibit recurring patterns (e.g., weekly or seasonal cycles) that influence the behavior of the series. Use both the observed values and the structure of the covariates to identify trends.
Predict the next \(H\) values based on the observed sequence and the upcoming covariate pattern. Just return the prediction values as a list. No explanation. \\
\midrule

\textbf{PromptCast~\citep{xue2023promptcast}} &
\ttfamily
From $\mathbf{x}_1$ to $\mathbf{x}_T$, there were \([y_1, y_2, y_3, \dots, y_T]\) values recorded. Predict the next $H$ values. Just return the prediction values as a list of numbers. No explanation.\\
\midrule

\textbf{Knowledge-Guided ~\citep{tang2025time}} &
\ttfamily
Same structure as the Decoupled format, but includes additional dataset or domain-specific information to guide forecasting (e.g., ``This time series represents energy demand in the U.S.''). \\
\bottomrule
\end{tabular}
\label{tab:prompt-examples}
\end{table}

\section{Experiments Setting}\label{sec:exp}
\subsection{Models and Datasets}
\begin{reviewtext}
We evaluate our approach with OpenAI’s GPT-4o-mini model on three real-world datasets from healthcare, service operations, and transportation, each widely used in time series prediction tasks \citep{zheng2025doubly, gruver2024large}. In all cases, covariates are extracted directly from the datasets, such as calendar date, month, or day-of-week, to capture temporal and seasonal patterns beyond the raw time series.
  
The first data set consists of weekly influenza positive cases obtained from the \href{https://www.who.int/tools/flunet}{WHO FluNet database} data set that spans the years 2016 to 2024. We focus on univariate time series corresponding to Influenza~A weekly positive cases in the United States, and the covariates include the calendar year, the calendar month (e.g., January or February), and a combined year-week indicator (e.g., 2024-W01), which are supposed to help encode temporal and seasonal structure. For evaluation, we used 25 weeks for validation and 25 weeks for testing within the year 2024, and assessed performance over 1-, 2-, and 5-week forecast horizons.

The second data set is \href{https://www.kaggle.com/datasets/cityofoakland/oakland-call-center-public-work-service-requests}{Oakland call center}, which contains the records of daily call arrival volumes from 2012 to 2014. We restrict attention to 30-minute interval call volumes between 8:00~AM and 5:00~PM, and for evaluation, we select a seven-week interval with three weeks for validation and two weeks for testing. The forecasting horizons include short-term (one day ahead) and mid-term (one week ahead) predictions, and the covariates are the exact date and the day of the week.

The third dataset is the Air Passengers time series, which reports monthly totals of international airline passengers from 1949 to 1960 (144 observations). We used the last four years (January~1956 to December~1959), with the first two years reserved for validation and the final two years for testing. The forecast horizons are 1, 6, and 12 months, and the covariates consist of the exact date and a categorical month-of-year indicator.

\end{reviewtext}

\subsection{Evaluation Metrics}

To evaluate predictive accuracy, this article reports three commonly used metrics: Mean Absolute Error (MAE), Mean Absolute Percentage Error (MAPE), and Root Mean Squared Error (RMSE).
The MAE measures the average magnitude of prediction errors without considering their direction:
\[
\mathrm{MAE} = \frac{1}{n} \sum_{i=1}^{n} |\hat{y}_i - y_i|,
\]
where $\hat{y}_i$ is the predicted value and $y_i$ is the ground truth at time point $i$.
The MAPE expresses this error as a percentage of the actual values, providing a scale-independent measure of accuracy:
\[
\mathrm{MAPE} = \frac{1}{n} \sum_{i=1}^{n} \left| \frac{\hat{y}_i - y_i}{y_i} \right|.
\]
While MAPE is widely used, we also calculate RMSE that penaltizes larger errors more heavily and preserves the original units of the target variable.
\[
\mathrm{RMSE} = \sqrt{ \frac{1}{n} \sum_{i=1}^{n} (\hat{y}_i - y_i)^2 }.
\]
Together, these metrics provide a comprehensive view of model performance.

\section{Numerical Results}\label{sec:res_n}
\subsection{Prompt–Covariate Pair Selection}
\begin{reviewtext}
As this step represents the first attempt to systematically leverage covariates for time series forecasting, a validation-based approach is adopted to select the best prompt–covariate pair for each dataset. Table~\ref{tab:all_datasets_results} reports the validation and test results in the data sets for three prompt designs: Coupled, Decoupled, and Contextualized.

Several key observations emerge. In about 85\% of the cases, the best prompt in the validation set also delivers the best performance on the test set. Moreover, the best prompt--covariate pairs identified in validation either remain the best or achieve accuracy comparable to the best on the test set, depending on the chosen metric. For instance, in the Call Center dataset, the Coupled prompt with Day of Week as a covariate achieves the best results in both validation and test sets. In the Influenza dataset with a forecast horizon of 2, the Coupled prompt with the Month covariate achieves the best performance across all metrics in validation, and on the test set, this pair yields the lowest RMSE, but not the lowest MAE or MAPE. 
 
We could also see that the Coupled prompt achieves strong and stable performance across datasets, generally outperforming both Decoupled and Contextualized alternatives by large margins.  
Another observation is that sometimes the optimal covariate depends on the forecasting horizon. For example, when predicting two weeks ahead in the Influenza dataset, the Month covariate yields the best results, whereas for a five-week forecast horizon, the Year covariate performs better. Taken together, these results confirm that the validation set provides a reliable guide for selecting prompt--covariate pairs. 

\end{reviewtext}
\begin{table}[H]
\centering
\caption{Forecasting results across datasets by horizon, covariate, and prompt design for validation and test sets.}
\label{tab:all_datasets_results}
\resizebox{\textwidth}{!}{%
\begin{tabular}{c|c|c|c|ccc|ccc}
\toprule
\textbf{Dataset} & \textbf{Forecast Horizon} & \textbf{Covariate} & \textbf{Prompt} & 
\multicolumn{3}{c|}{\textbf{Validation}} & 
\multicolumn{3}{c}{\textbf{Test}} \\
\cline{5-10}
& & & & \textbf{RMSE} & \textbf{MAE} & \textbf{MAPE} (\%)& \textbf{RMSE} & \textbf{MAE} & \textbf{MAPE} (\%)\\
\midrule
\multirow{27}{*}{Influenza}
 & \multirow{9}{*}{1}
 & \multirow{3}{*}{Year}
 & Coupled        & 564.18   & 315.16   & 8.29   & 1964.97  & 797.32   & 15.11 \\
 & & & Decoupled      & 12841.83 & 4985.04  & 476.44 & 17494.99 & 9051.56  & 1615.06 \\
 & & & Contextualized & 4352.89  & 3233.92  & 288.48 & 12008.47 & 5343.28  & 1071.19 \\
 \cline{3-10}
 & & \multirow{3}{*}{Month}
 & Coupled        & 768.26   & 505.08   & 10.38  & 1682.17  & 592.28   & 13.78 \\
 & & & Decoupled      & 10888.62 & 5414.68  & 377.94 & 16262.22 & 8857.16  & 1222.88 \\
 & & & Contextualized & 15614.99 & 8496.88  & 628.22 & 13111.15 & 8963.92  & 1470.28 \\
 \cline{3-10}
 & & \multirow{3}{*}{Date}
 & Coupled        & 1084.00  & 556.48   & 9.52   & 1343.32  & 540.28   & 13.70 \\
 & & & Decoupled      & 2912.15  & 1729.63  & 151.87 & 9107.15  & 3010.60  & 460.45 \\
 & & & Contextualized & 23521.77 & 14234.68 & 1681.51 & 32298.58 & 23936.16 & 4188.99 \\
\cline{2-10}
 & \multirow{9}{*}{2}
 & \multirow{3}{*}{Year}
 & Coupled        & 1179.99  & 635.04   & 12.92  & 1969.99  & 808.36   & 17.84 \\
 & & & Decoupled      & 2919.09  & 2140.24  & 121.35 & 3463.82  & 1806.20  & 150.00 \\
 & & & Contextualized & 7990.04  & 4234.16  & 429.99 & 17626.56 & 9044.36  & 2053.65 \\
 \cline{3-10}
 & & \multirow{3}{*}{Month}
 & Coupled        & 738.45   & 438.24   & 9.52   & 1724.58  & 1238.56  & 30.36 \\
 & & & Decoupled      & 16424.49 & 7586.64  & 711.53 & 7239.28  & 4792.96  & 503.03 \\
 & & & Contextualized & 4232.21  & 3146.80  & 176.02 & 8127.52  & 7007.72  & 1115.88 \\
 \cline{3-10}
 & & \multirow{3}{*}{Date}
 & Coupled        & 1075.28  & 603.72   & 10.00  & 2976.31  & 757.52   & 18.72 \\
 & & & Decoupled      & 23783.87 & 13662.32 & 1867.49 & 12413.28 & 4586.12  & 563.93 \\
 & & & Contextualized & 22495.99 & 10966.08 & 1433.17 & 34314.72 & 23307.08 & 3630.98 \\
\cline{2-10}
 & \multirow{9}{*}{5}
 & \multirow{3}{*}{Year}
 & Coupled        & 1193.08  & 789.48   & 18.47  & 2250.02  & 1573.40  & 32.27 \\
 & & & Decoupled      & 4740.10  & 3638.40  & 286.19 & 6014.52  & 2895.80  & 95.27 \\
 & & & Contextualized & 6793.82  & 5111.72  & 447.85 & 6720.57  & 5233.00  & 751.63 \\
 \cline{3-10}
 & & \multirow{3}{*}{Month}
 & Coupled        & 1374.72  & 792.20   & 13.85  & 3805.09  & 1197.32  & 39.67 \\
 & & & Decoupled      & 3975.82  & 3029.48  & 267.34 & 14610.38 & 9575.44  & 1070.37 \\
 & & & Contextualized & 6835.68  & 5371.36  & 525.50 & 8875.65  & 7188.44  & 946.63 \\
 \cline{3-10}
 & & \multirow{3}{*}{Date}
 & Coupled        & 1407.35  & 819.56   & 18.86  & 6190.20  & 2579.56  & 41.41 \\
 & & & Decoupled      & 5045.80  & 3490.08  & 89.22  & 37948.85 & 21213.24 & 1099.31 \\
 & & & Contextualized & 19709.42 & 15839.20 & 1645.34 & 50079.06 & 45460.72 & 8482.24 \\
\midrule
\multirow{12}{*}{Call Center}
 & \multirow{6}{*}{1}
 & \multirow{3}{*}{Date}
 & Coupled        & 155.62 & 116.19 & 33.38 & 152.63 & 107.50 & 29.66 \\
 & & & Decoupled      & 175.54 & 130.48 & 40.07 & 166.90 & 115.43 & 33.21 \\
 & & & Contextualized & 173.12 & 127.38 & 39.41 & 167.26 & 122.54 & 37.53 \\
 \cline{3-10}
 & & \multirow{3}{*}{Day of Week}
 & Coupled        & 115.97 & 75.76  & 17.04 & 91.19  & 56.18  & 14.69 \\
 & & & Decoupled      & 193.90 & 144.90 & 42.79 & 192.50 & 147.29 & 42.83 \\
 & & & Contextualized & 179.74 & 136.14 & 41.22 & 177.56 & 133.89 & 39.38 \\
\cline{2-10}
 & \multirow{6}{*}{7}
 & \multirow{3}{*}{Date}
 & Coupled        & 196.18 & 157.05 & 42.51 & 178.10 & 128.54 & 35.90 \\
 & & & Decoupled      & 187.66 & 135.52 & 42.26 & 178.26 & 131.25 & 40.25 \\
 & & & Contextualized & 235.56 & 175.43 & 53.71 & 177.71 & 130.96 & 40.09 \\
 \cline{3-10}
 & & \multirow{3}{*}{Day of Week}
 & Coupled        & 109.65 & 72.24  & 15.40 & 86.56  & 51.64  & 12.93 \\
 & & & Decoupled      & 209.13 & 151.62 & 47.27 & 200.39 & 152.71 & 45.53 \\
 & & & Contextualized & 213.20 & 153.38 & 47.63 & 190.65 & 136.57 & 42.41 \\
\midrule
\multirow{18}{*}{Air Passengers}
 & \multirow{6}{*}{1}
 & \multirow{3}{*}{Date}
 & Coupled        & 39.22 & 30.21 & 8.39 & 41.47 & 31.25 & 6.78 \\
 & & & Decoupled      & 88.25 & 54.38 & 15.02 & 84.56 & 62.96 & 14.58 \\
 & & & Contextualized & 77.91 & 52.92 & 14.40 & 88.94 & 68.13 & 16.11 \\
 \cline{3-10}
 & & \multirow{3}{*}{Month}
 & Coupled        & 50.26 & 37.21 & 10.48 & 43.74 & 32.08 & 7.34 \\
 & & & Decoupled      & 46.74 & 37.92 & 10.04 & 52.72 & 43.71 & 9.71 \\
 & & & Contextualized & 48.09 & 37.83 & 10.03 & 55.93 & 46.88 & 10.39 \\
\cline{2-10}
 & \multirow{6}{*}{6}
 & \multirow{3}{*}{Date}
 & Coupled        & 47.45 & 41.88 & 11.50 & 55.95 & 45.33 & 9.75 \\
 & & & Decoupled      & 126.00 & 95.79 & 26.54 & 88.62 & 73.13 & 16.00 \\
 & & & Contextualized & 75.85 & 62.46 & 16.88 & 94.42 & 68.04 & 14.30 \\
 \cline{3-10}
 & & \multirow{3}{*}{Month}
 & Coupled        & 29.47 & 26.92 & 7.42 & 70.90 & 56.50 & 12.72 \\
 & & & Decoupled      & 81.07 & 70.79 & 19.82 & 68.85 & 54.42 & 12.39 \\
 & & & Contextualized & 59.67 & 46.42 & 11.77 & 48.83 & 35.42 & 7.43 \\
\cline{2-10}
 & \multirow{6}{*}{12}
 & \multirow{3}{*}{Date}
 & Coupled        & 27.07 & 20.08 & 5.28 & 31.26 & 25.08 & 5.63 \\
 & & & Decoupled      & 56.77 & 42.79 & 11.40 & 123.69 & 94.67 & 19.62 \\
 & & & Contextualized & 86.14 & 63.71 & 18.18 & 98.84 & 85.71 & 19.46 \\
 \cline{3-10}
 & & \multirow{3}{*}{Month}
 & Coupled        & 19.52 & 14.96 & 3.86 & 33.84 & 30.54 & 6.68 \\
 & & & Decoupled      & 76.00 & 59.67 & 16.33 & 73.18 & 57.79 & 12.98 \\
 & & & Contextualized & 60.67 & 50.50 & 13.18 & 82.75 & 68.54 & 15.13 \\
\bottomrule
\end{tabular}
}
\end{table}
\vspace{1em}

\subsection{Comparison with No-Covariate Baseline}
\begin{reviewtext}
We next compared the best-performing prompt with the results of the No-Covariate prompt, which represents the simplest form of time series forecasting with LLMs. 
Table~\ref{tab:all_datasets_coupled_covariates} reports the results across all datasets and forecast horizons. 
For the Influenza dataset, incorporating covariate reduces RMSE by 76\%, 53\%, and 73\% in the forecast horizons 1, 2 and 5, respectively. 
In the Call Center dataset, covariates lead to error reductions of 47\% for the 1-day forecast horizon and 54\% for the 7-day forecast horizon. 
For Air Passengers, the gains are equally substantial, with decreases of 54\%, 42\%, and 70\% over forecast horizons of 1, 6, and 12 months, respectively. 
Although covariate integration consistently improves over the No-Covariate prompt, its amount depends on the chosen covariate. 
For example, in the Call Center dataset with a 7-day horizon, selecting Date as the covariate produces performance close to the No-Covariate prompt, which highlights the importance of careful covariate selection.

\end{reviewtext}

\begin{table}[H]
\centering
\caption{Forecasting results of the Coupled prompt versus the No-Covariate prompt across datasets, forecast horizons, and covariates for both validation and test sets.}
\label{tab:all_datasets_coupled_covariates}
\resizebox{\textwidth}{!}{%
\begin{tabular}{c|c|c|ccc|ccc}
\toprule
\textbf{Dataset} & \textbf{Forecast Horizon} & \textbf{Covariate} &
\multicolumn{3}{c|}{\textbf{Validation}} &
\multicolumn{3}{c}{\textbf{Test}} \\
\cline{4-9}
& & & \textbf{RMSE} & \textbf{MAE} & \textbf{MAPE} (\%) & \textbf{RMSE} & \textbf{MAE} & \textbf{MAPE} (\%) \\
\midrule
\multirow{12}{*}{Influenza}
  & \multirow{4}{*}{1}
  & No--Covariate & 1385.61 & 859.28  & 23.10  & 5673.10 & 1703.80 & 44.82 \\
  & & Year         & 564.18  & 315.16  & 8.29   & 1964.97 & 797.32  & 15.11 \\
  & & Month        & 768.26  & 505.08  & 10.38  & 1682.17 & 592.28  & 13.78 \\
  & & Date         & 1084.00 & 556.48  & 9.52   & 1343.32 & 540.28  & 13.70 \\
\cline{2-9}
  & \multirow{4}{*}{2}
  & No--Covariate & 9822.05 & 3546.88 & 128.86 & 3404.30 & 1269.20 & 23.22 \\
  & & Year         & 1180.00 & 635.04  & 12.92  & 1969.99 & 808.36  & 17.84 \\
  & & Month        & 738.45  & 438.24  & 9.52   & 1724.58 & 1238.56 & 30.36 \\
  & & Date         & 1075.28 & 603.72  & 10.00  & 2976.31 & 757.52  & 18.72 \\
\cline{2-9}
  & \multirow{4}{*}{5}
  & No--Covariate & 7057.82 & 4350.56 & 520.75 & 8530.29 & 3411.64 & 78.25 \\
  & & Year         & 1193.08 & 789.48  & 18.47  & 2250.02 & 1573.40 & 32.27 \\
  & & Month        & 1374.72 & 792.20  & 13.85  & 3805.09 & 1197.32 & 39.67 \\
  & & Date         & 1407.35 & 819.56  & 18.86  & 6190.20 & 2579.56  & 41.41 \\
\midrule
\multirow{6}{*}{Call Center}
  & \multirow{3}{*}{1}
  & No--Covariate   & 173.12 & 124.67 & 38.85 & 172.54 & 132.93 & 39.40 \\
  & & Date           & 155.62 & 116.19 & 33.38 & 152.63 & 107.50 & 29.66 \\
  & & Day of Week    & 115.97 & 75.76  & 17.04 & 91.19  & 56.18  & 14.69 \\
\cline{2-9}
  & \multirow{3}{*}{7}
  & No--Covariate   & 148.03 & 108.14 & 31.49 & 187.63 & 144.29 & 43.08 \\
  & & Date           & 196.18 & 157.05 & 42.51 & 178.10 & 128.54 & 35.90 \\
  & & Day of Week    & 109.65 & 72.24  & 15.40 & 86.56  & 51.64  & 12.93 \\
\midrule
\multirow{9}{*}{Air Passengers}
  & \multirow{3}{*}{1}
  & No--Covariate  & 64.09 & 50.33 & 14.12 & 91.65 & 63.54 & 14.87 \\
  & & Date                    & 39.22 & 30.21 &  8.39 & 41.47 & 31.25 &  6.78 \\
  & & Month                   & 50.26 & 37.21 & 10.48 & 43.74 & 32.08 &  7.34 \\
\cline{2-9}
  & \multirow{3}{*}{6}
  & No--Covariate  & 80.24 & 66.42 & 18.33 & 96.93 & 79.38 & 18.10 \\
  & & Date                    & 47.45 & 41.88 & 11.50 & 55.95 & 45.33 &  9.75 \\
  & & Month                   & 29.47 & 26.92 &  7.42 & 70.90 & 56.50 & 12.72 \\
\cline{2-9}
  & \multirow{3}{*}{12}
  & No--Covariate  & 85.12 & 66.21 & 18.81 & 106.84 & 77.83 & 17.52 \\
  & & Date                    & 27.07 & 20.08 &  5.28 & 31.26 & 25.08 &  5.63 \\
  & & Month                   & 19.52 & 14.96 &  3.86 & 33.84 & 30.54 &  6.68 \\
\bottomrule
\end{tabular}
}
\end{table}

\begin{figure}[H]
\centering
\begin{subfigure}{0.45\textwidth}
  \includegraphics[width=\linewidth]{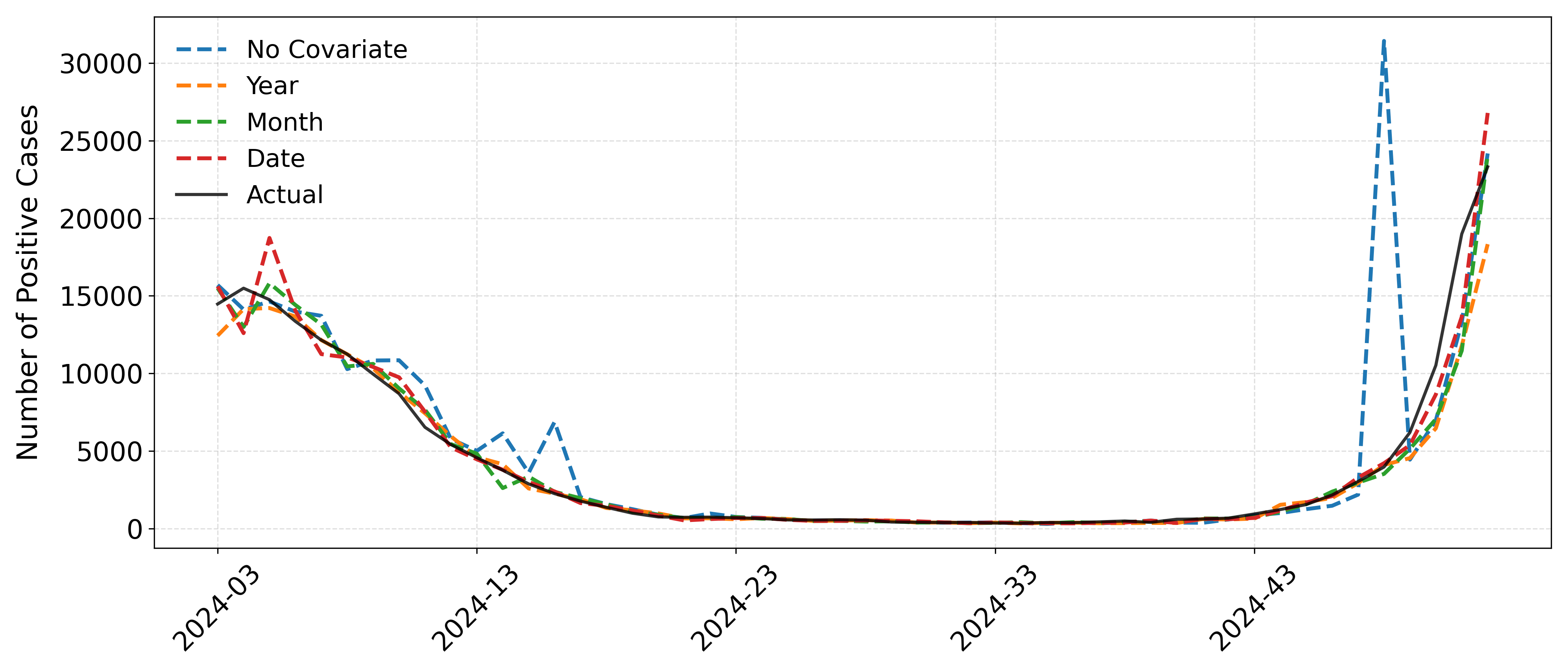}
  \caption{Influenza (1 week)}
\end{subfigure}
\begin{subfigure}{0.45\textwidth}
  \includegraphics[width=\linewidth]{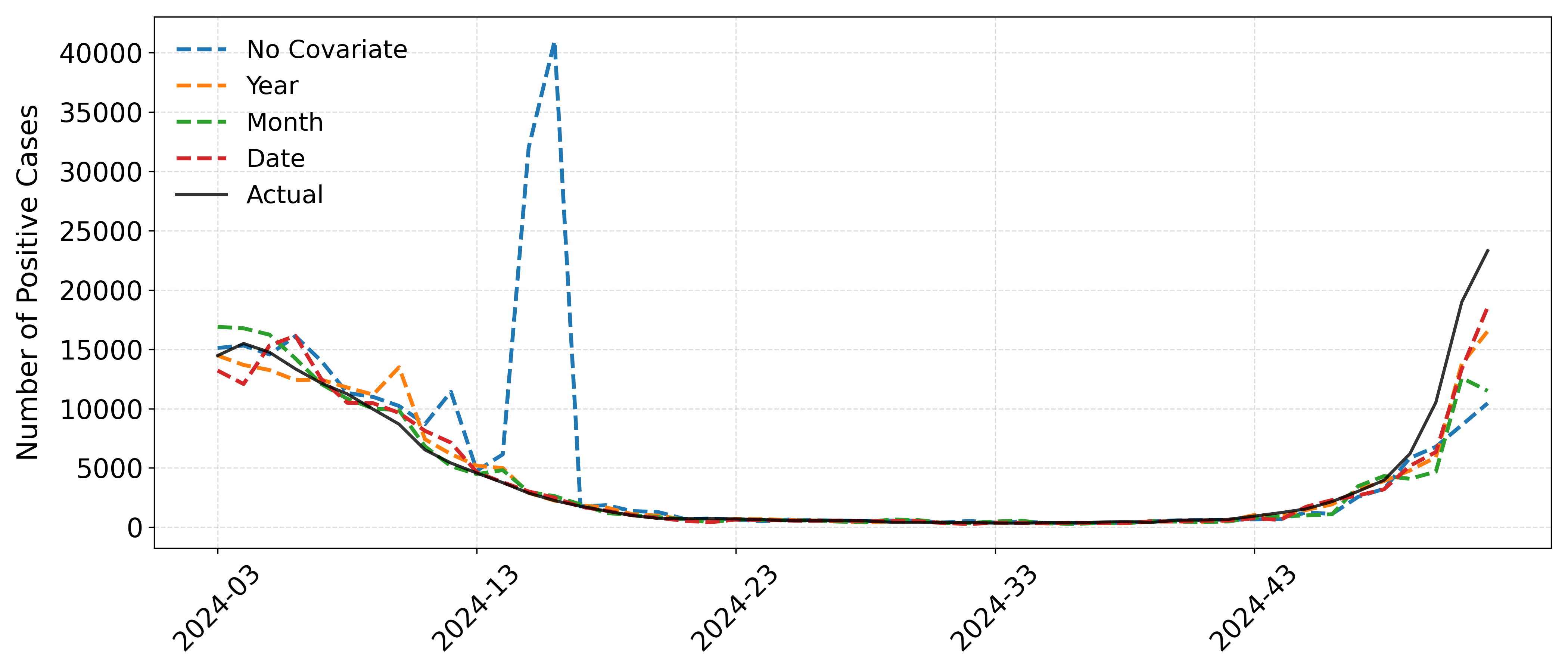}
  \caption{Influenza (2 weeks)}
\end{subfigure}

\begin{subfigure}{0.45\textwidth}
  \includegraphics[width=\linewidth]{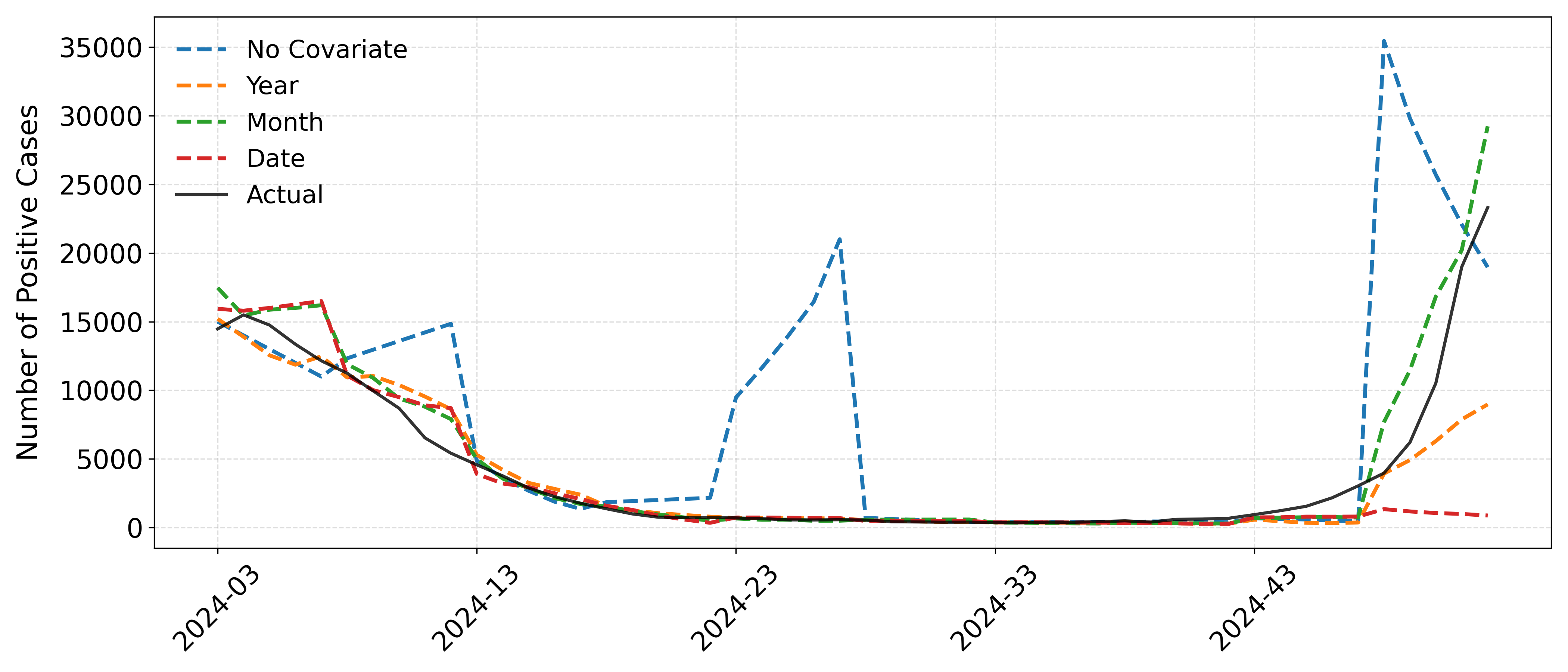}
  \caption{Influenza (5 weeks)}
\end{subfigure}
\begin{subfigure}{0.45\textwidth}
  \includegraphics[width=\linewidth]{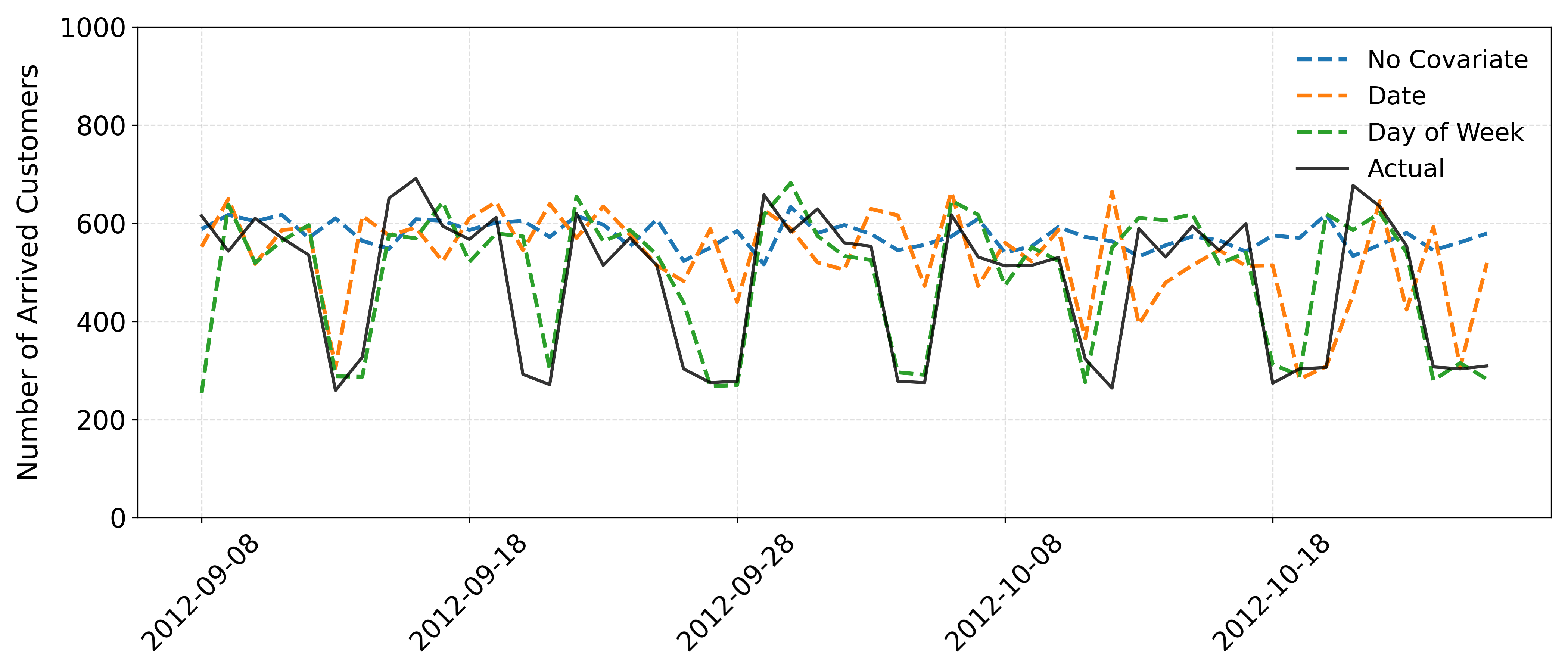}
  \caption{Call Center (1 day)}
\end{subfigure}

\begin{subfigure}{0.45\textwidth}
  \includegraphics[width=\linewidth]{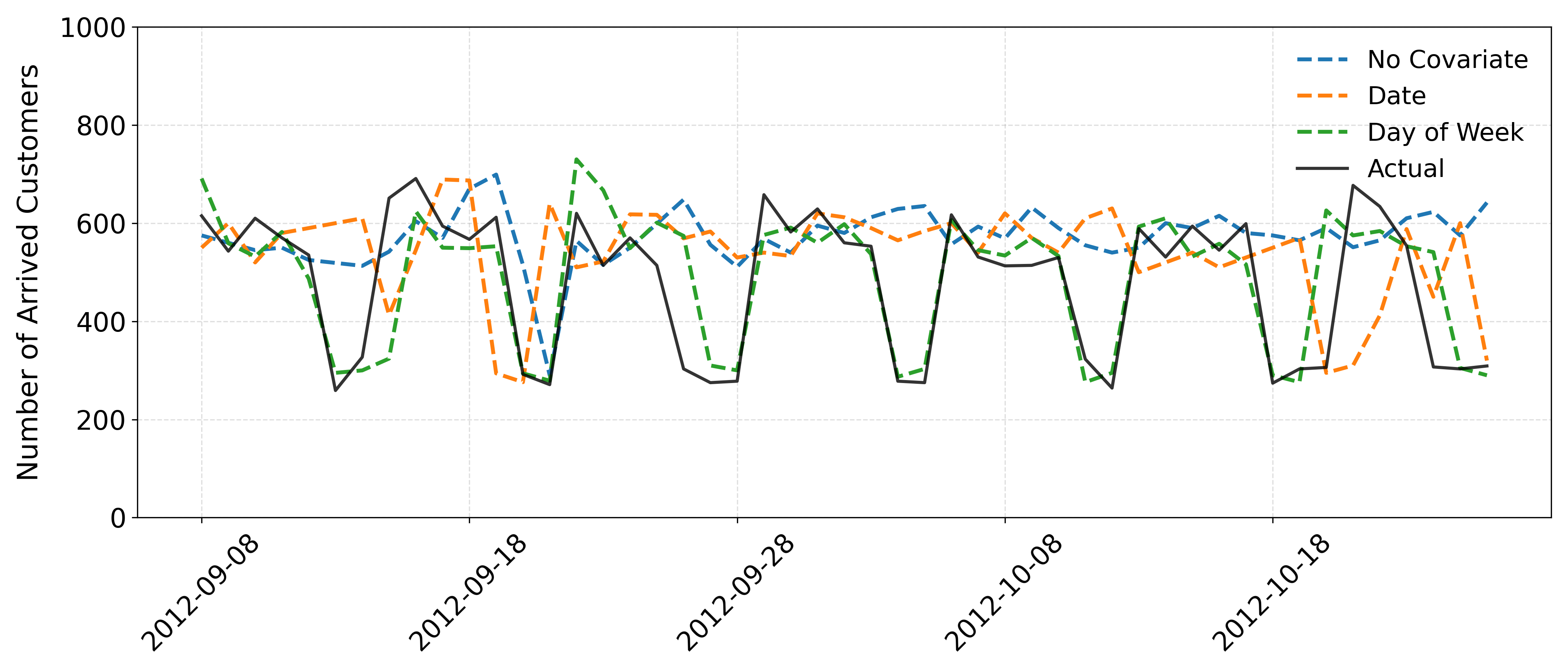}
  \caption{Call Center (7 days)}
\end{subfigure}
\begin{subfigure}{0.45\textwidth}
  \includegraphics[width=\linewidth]{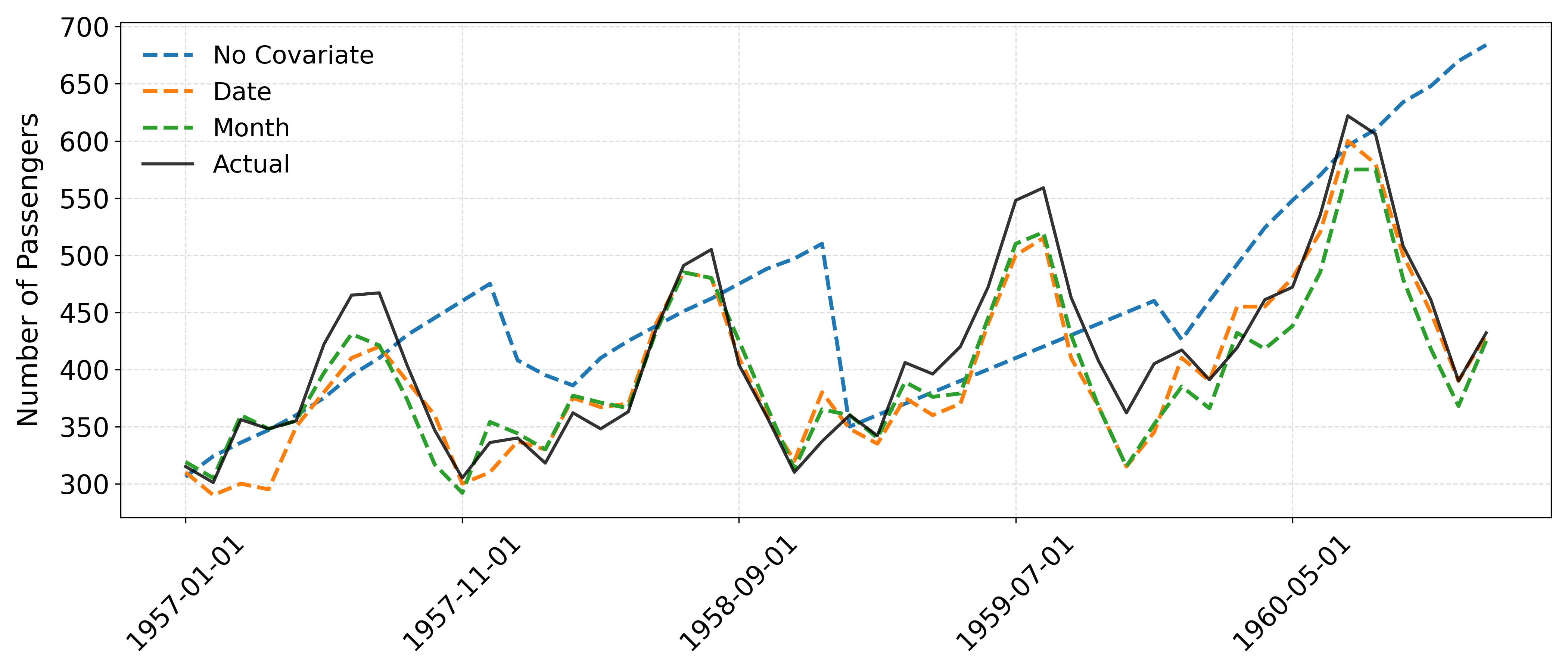}
  \caption{Air Passengers (1 month)}
\end{subfigure}

\begin{subfigure}{0.45\textwidth}
  \includegraphics[width=\linewidth]{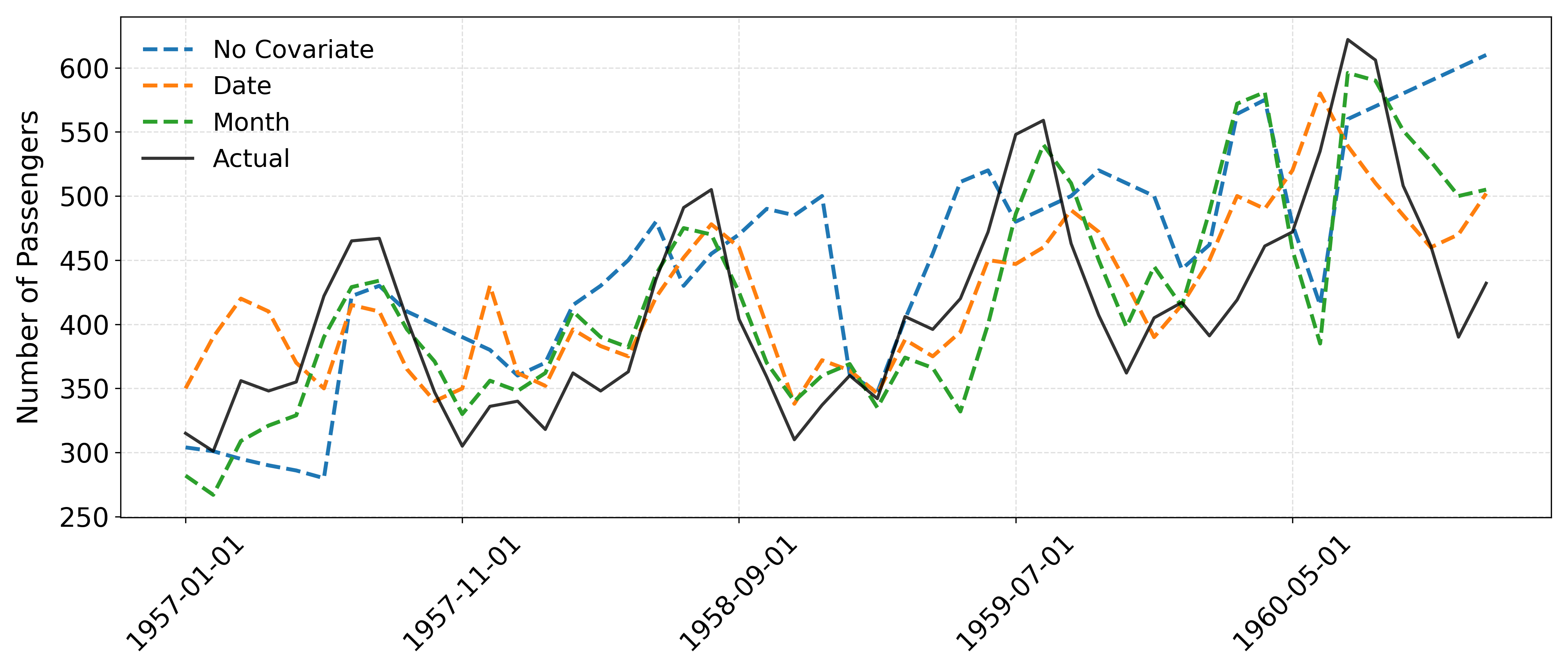}
  \caption{Air Passengers (6 months)}
\end{subfigure}
\begin{subfigure}{0.45\textwidth}
  \includegraphics[width=\linewidth]{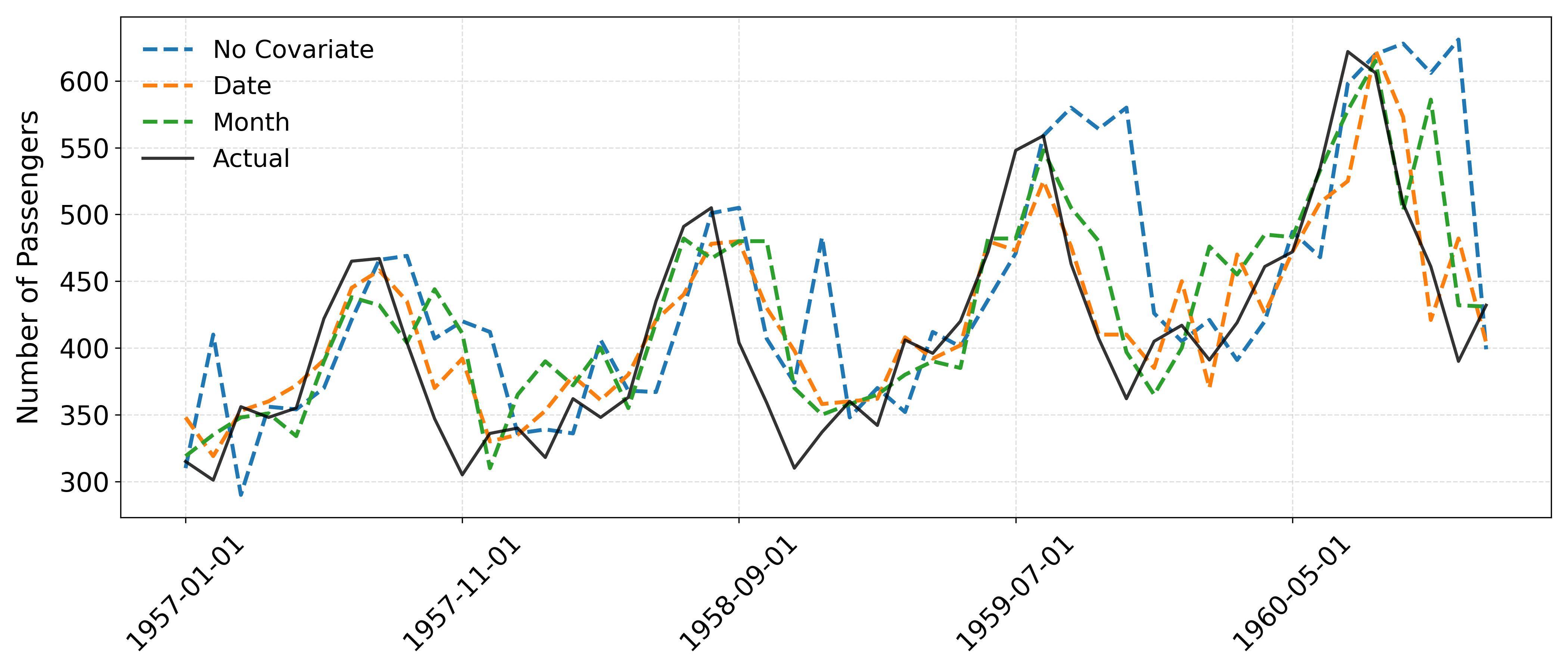}
  \caption{Air Passengers (12 months)}
\end{subfigure}

\caption{Forecasts of the Coupled prompt with covariate integration versus the No-Covariate prompt across datasets, forecast horizons, and covariates for both validation and test sets.}
\label{fig:forecasting_results}
\end{figure}

\subsection{Comparison with Existing Prompting Strategies}
\begin{reviewtext}
To assess the comparative effectiveness of our approach, we benchmark the Coupled prompt against PromptCast and Knowledge-Guided prompt with the results summarized in Table~\ref{tab:all_datasets_prompt_comparison}. The results show that Coupled prompt consistently 
matches or outperforms existing methods and offers both a lower average error and
greater stability.  

For the Influenza dataset, coupled prompt achieves substantially lower error across all
forecast horizons. PromptCast provides modest gains in a few cases but lacks
consistency, while Knowledge-Guided prompting often performs poorly, especially
at longer horizons where its errors grow dramatically.  
For the Call Center dataset, both PromptCast and Knowledge-Guided lag behind, and the reductions in RMSE are similar across forecast horizons, averaging around 50\%.
For the Air Passengers dataset, the difference at the 1-month forecast horizon is negligible, but at longer forecast horizons, the Coupled prompt achieves clear gains, reducing RMSE by at least 16\% at 6 months and 59\% at 12 months. 
Overall, the Coupled prompt shows superior performance across datasets and forecast horizons and yields more accurate forecasts, with the advantage more evident in longer-term predictions.

\end{reviewtext}
\begin{table}[H]
\centering
\caption{Forecasting results using Coupled prompt compared to PromptCast and Knowledge-Guided prompt across datasets and forecast horizons.}
\label{tab:all_datasets_prompt_comparison}
\resizebox{\textwidth}{!}{%
\begin{tabular}{l l l rrr rrr}
\toprule
\textbf{Dataset} & \textbf{Forecast Horizon} & \textbf{Prompt} &
\multicolumn{3}{c}{\textbf{Validation}} &
\multicolumn{3}{c}{\textbf{Test}} \\
\cmidrule(lr){4-6} \cmidrule(lr){7-9}
& & & \textbf{RMSE} & \textbf{MAE} & \textbf{MAPE} (\%) & \textbf{RMSE} & \textbf{MAE} & \textbf{MAPE} (\%) \\
\midrule
\multirow{9}{*}{Influenza}
& \multirow{3}{*}{1}
  & Coupled          & 1084.00 & 556.48 &  9.52 & 1343.32 &  540.28 &  13.70 \\
& & PromptCast       & 1866.47 & 872.52 & 57.91 & 2685.95 & 1073.36 &  18.12 \\
& & Knowledge-Guided & 2767.88 & 1731.24&140.50 & 5486.29 & 2358.44 & 249.49 \\
\cmidrule(l){2-9}
& \multirow{3}{*}{2}
  & Coupled          &  738.45 & 438.24 &  9.52 & 1724.58 & 1238.56 &  30.36 \\
& & PromptCast       & 5217.58 & 2148.36&108.55 & 3282.67 & 1384.28 &  26.70 \\
& & Knowledge-Guided &11645.30 & 5123.48&226.25 & 3566.82 & 1829.88 &  63.00 \\
\cmidrule(l){2-9}
& \multirow{3}{*}{5}
  & Coupled          & 1374.72 & 792.20 & 13.85 & 3805.09 & 1197.32 &  39.67 \\
& & PromptCast       & 4198.49 & 2398.32& 70.41 & 5223.67 & 2237.40 &  48.57 \\
& & Knowledge-Guided &32420.12 &15613.40&2448.87& 6017.91 & 2443.12 &  42.23 \\
\midrule
\multirow{6}{*}{Call Center}
& \multirow{3}{*}{1}
  & Coupled          &  115.97 &  75.76 & 17.04 &   91.19 &   56.18 &  14.69 \\
& & PromptCast       &  152.50 & 110.29 & 34.19 &  173.95 &  129.89 &  39.22 \\
& & Knowledge-Guided &  154.76 & 116.29 & 35.35 &  202.57 &  154.68 &  40.05 \\
\cmidrule(l){2-9}
& \multirow{3}{*}{7}
  & Coupled          &  109.65 &  72.24 & 15.40 &   86.56 &   51.64 &  12.93 \\
& & PromptCast       &  175.20 & 127.57 & 38.19 &  178.25 &  135.43 &  40.62 \\
& & Knowledge-Guided &  191.45 & 137.71 & 40.10 &  171.42 &  127.36 &  38.60 \\
\midrule
\multirow{9}{*}{Air Passengers}
& \multirow{3}{*}{1}
  & Coupled          &   50.26 &  37.21 & 10.48 &   43.74 &   32.08 &   7.34 \\
& & PromptCast       &   46.74 &  37.17 &  9.99 &   69.85 &   55.13 &  12.40 \\
& & Knowledge-Guided &   50.31 &  43.04 & 11.37 &   61.83 &   51.13 &  11.37 \\
\cmidrule(l){2-9}
& \multirow{3}{*}{6}
  & Coupled          &   29.47 &  26.92 &  7.42 &   70.90 &   56.50 &  12.72 \\
& & PromptCast       &   56.80 &  47.29 & 13.15 &   90.87 &   74.71 &  16.71 \\
& & Knowledge-Guided &   73.92 &  60.08 & 16.27 &   59.69 &   48.33 &  10.06 \\
\cmidrule(l){2-9}
& \multirow{3}{*}{12}
  & Coupled          &   19.52 &  14.96 &  3.86 &   33.84 &   30.54 &   6.68 \\
& & PromptCast       &   63.26 &  48.75 & 12.85 &   80.42 &   64.13 &  13.38 \\
& & Knowledge-Guided &   98.47 &  76.04 & 19.42 &   95.81 &   76.21 &  15.82 \\
\bottomrule
\end{tabular}
}
\end{table}

\begin{figure}[H]
\centering
\begin{subfigure}{0.45\textwidth}
  \includegraphics[width=\linewidth]{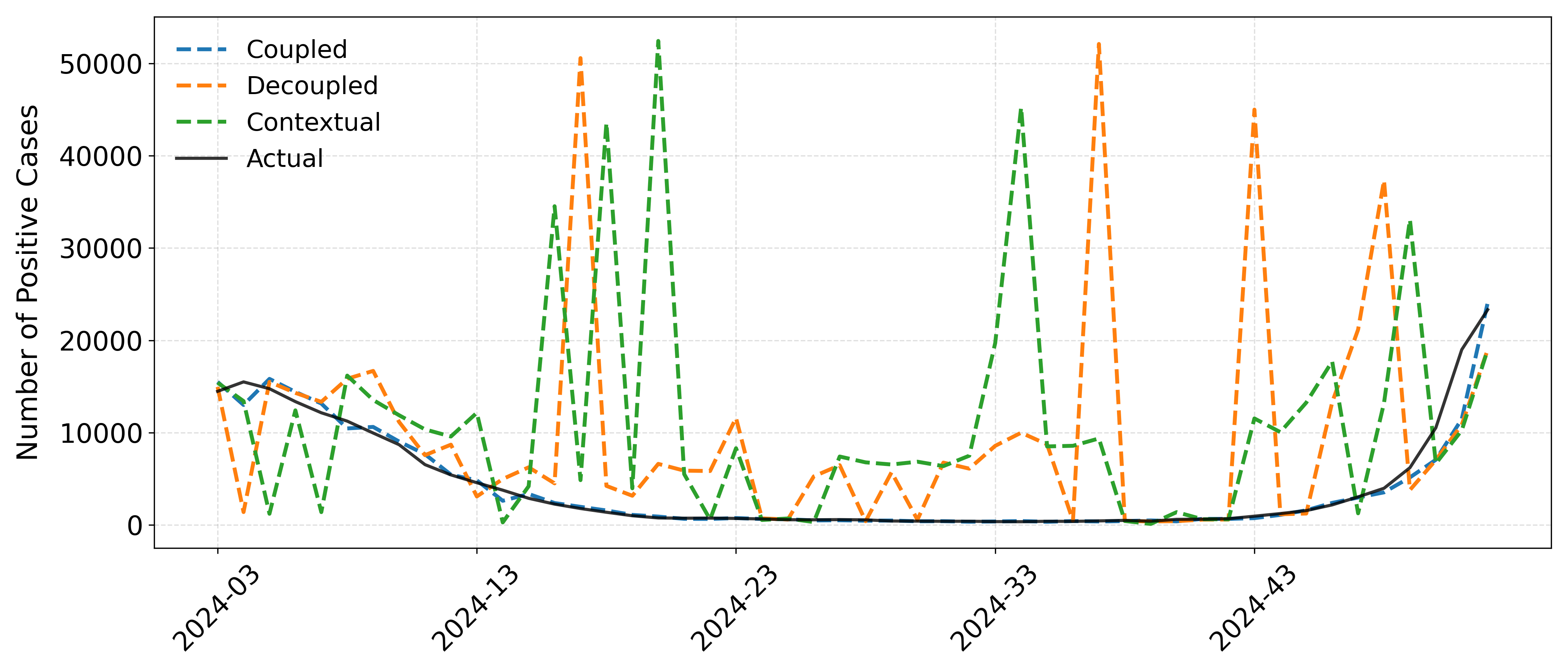}
  \caption{Influenza (1 week)}
\end{subfigure}
\begin{subfigure}{0.45\textwidth}
  \includegraphics[width=\linewidth]{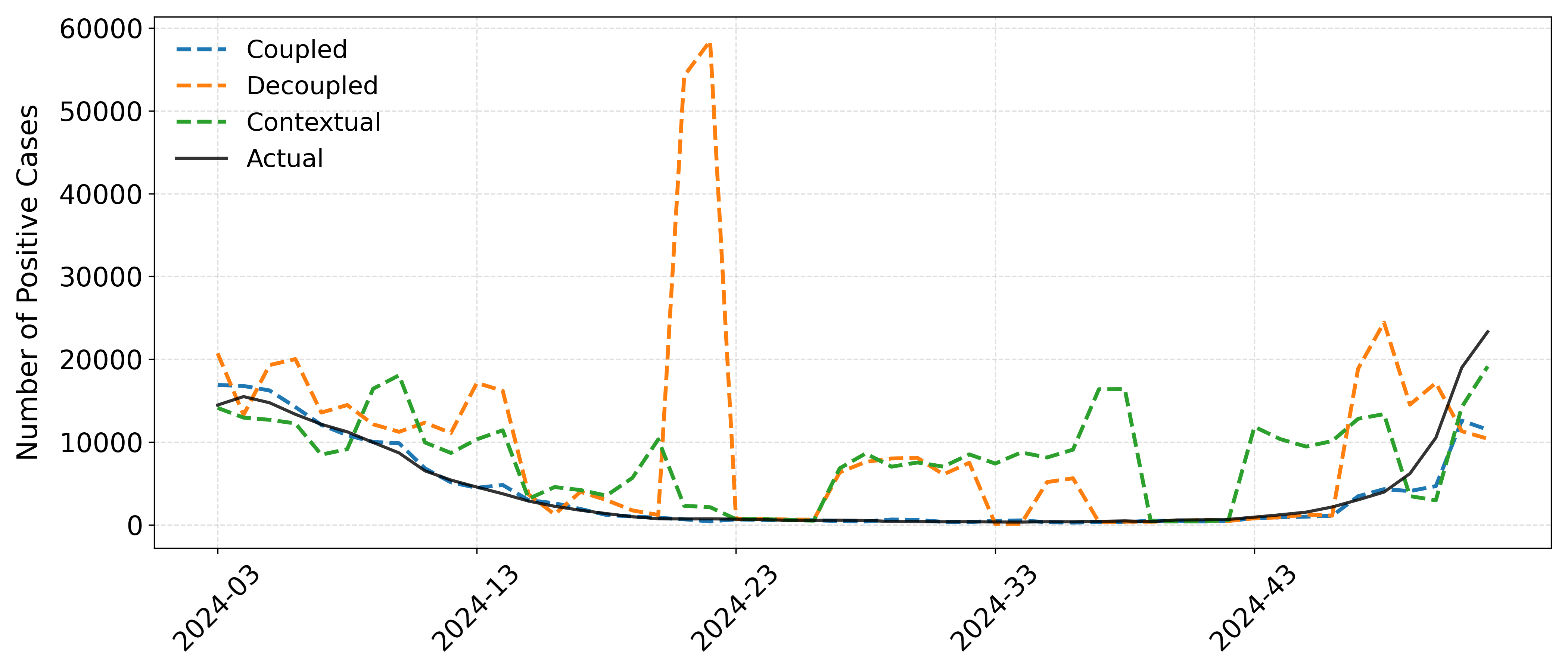}
  \caption{Influenza (2 weeks)}
\end{subfigure}

\begin{subfigure}{0.45\textwidth}
  \includegraphics[width=\linewidth]{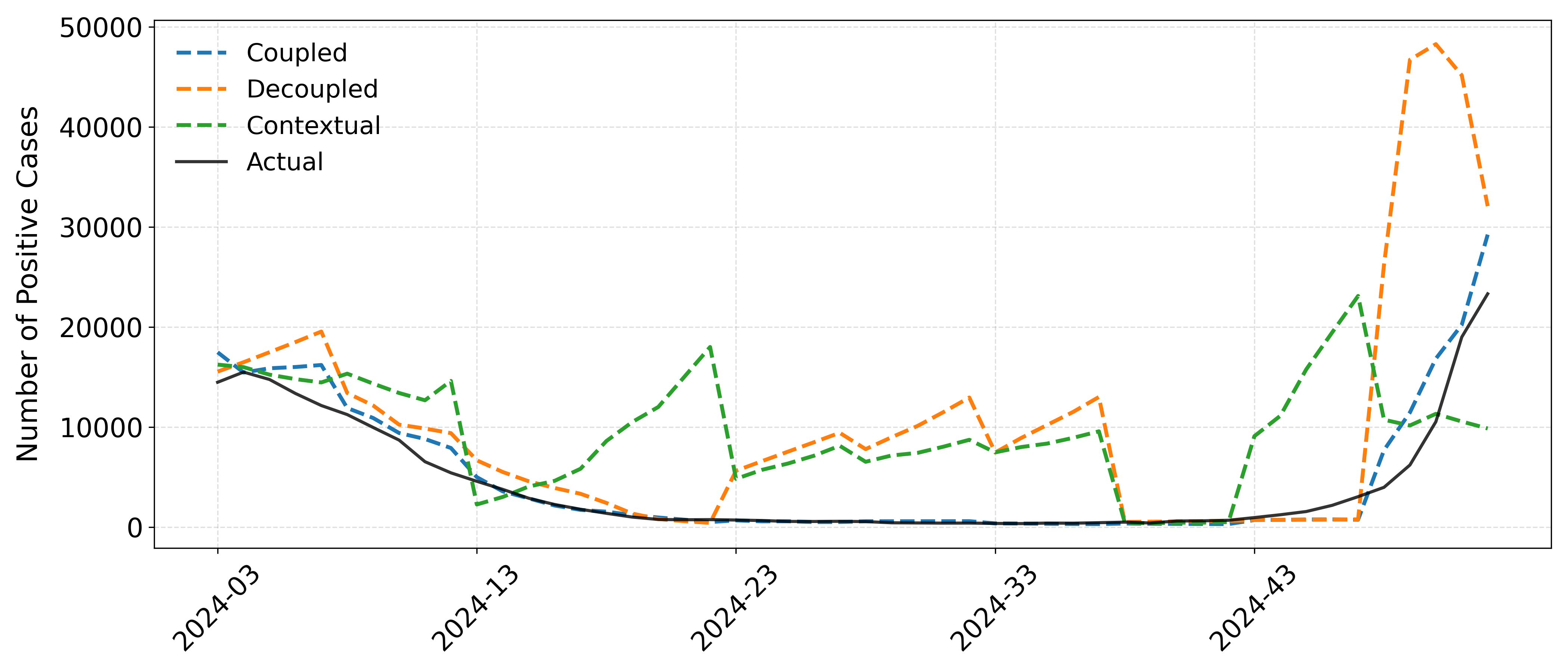}
  \caption{Influenza (5 weeks)}
\end{subfigure}
\begin{subfigure}{0.45\textwidth}
  \includegraphics[width=\linewidth]{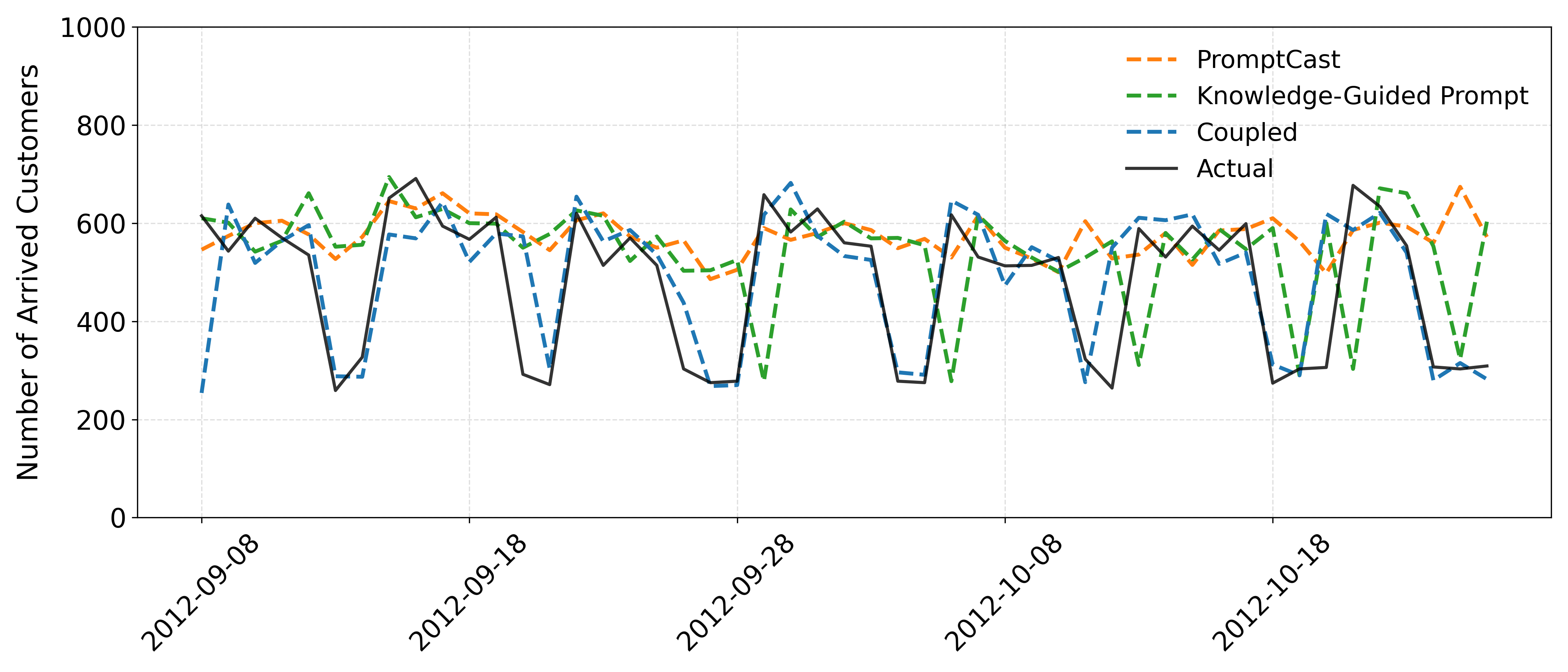}
  \caption{Call Center (1 day)}
\end{subfigure}

\begin{subfigure}{0.45\textwidth}
  \includegraphics[width=\linewidth]{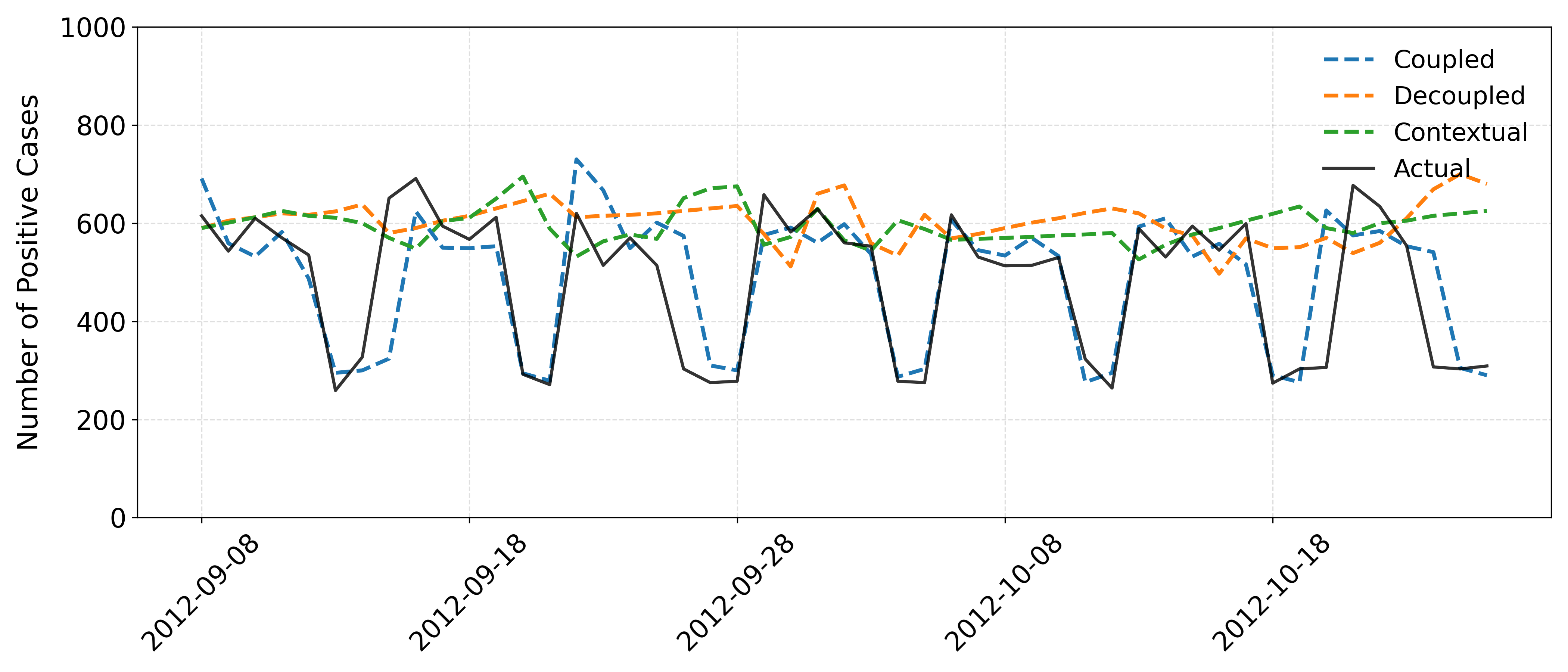}
  \caption{Call Center (7 days)}
\end{subfigure}
\begin{subfigure}{0.45\textwidth}
  \includegraphics[width=\linewidth]{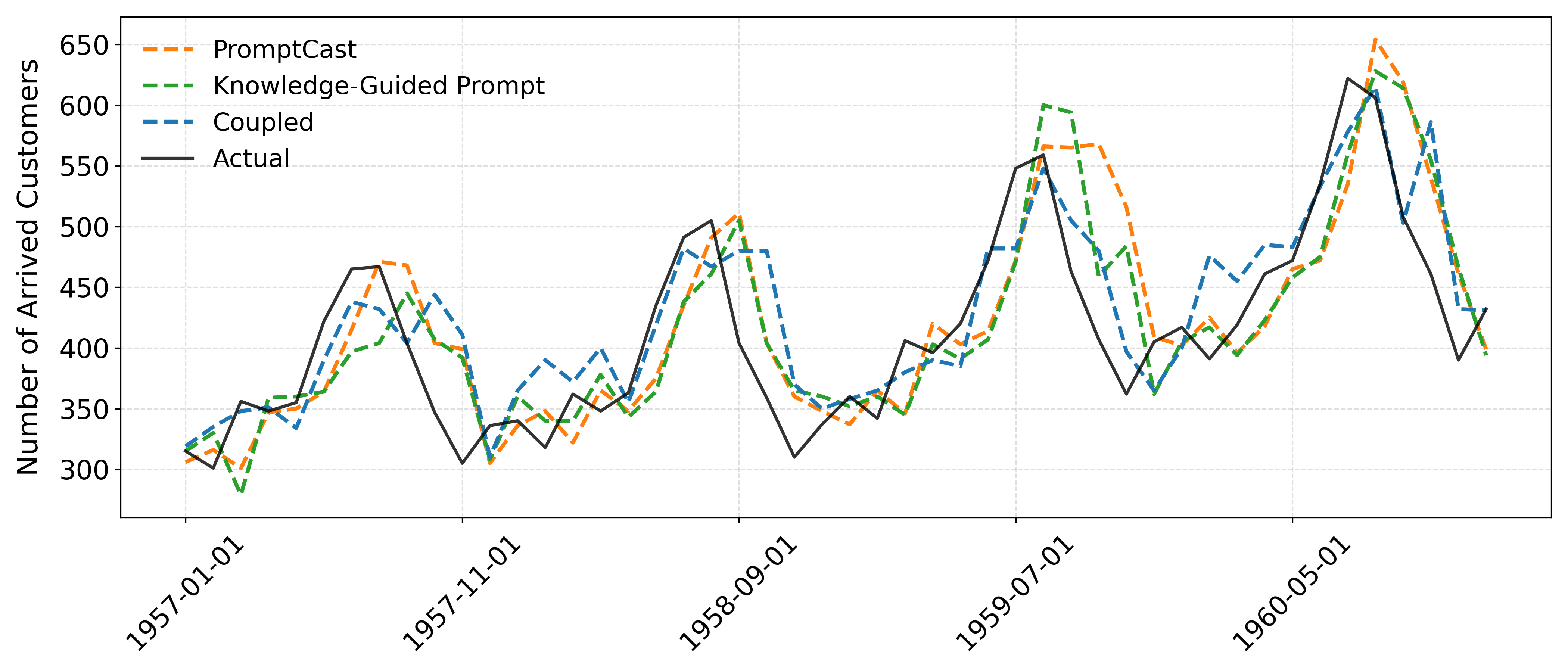}
  \caption{Air Passengers (1 month)}
\end{subfigure}

\begin{subfigure}{0.45\textwidth}
  \includegraphics[width=\linewidth]{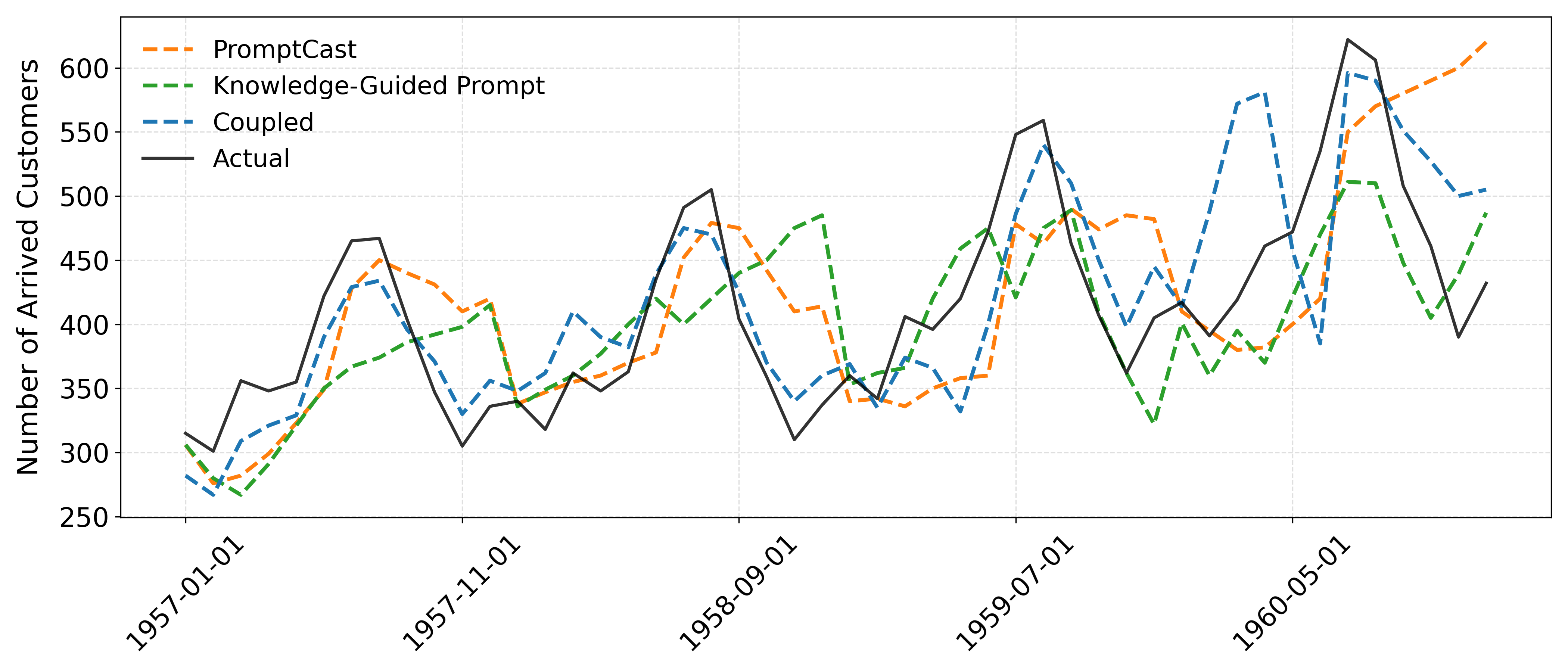}
  \caption{Air Passengers (6 months)}
\end{subfigure}
\begin{subfigure}{0.45\textwidth}
  \includegraphics[width=\linewidth]{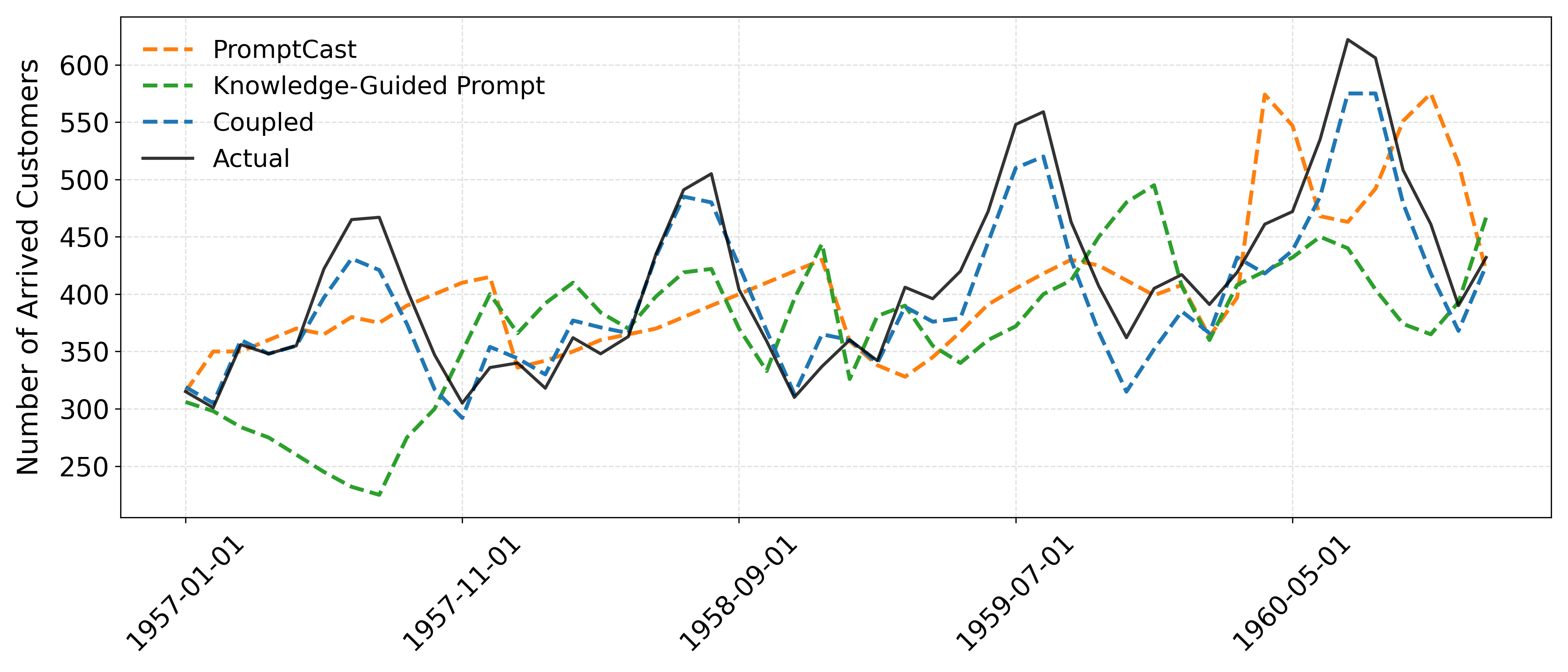}
  \caption{Air Passengers (12 months)}
\end{subfigure}

\caption{Forecasts of the Coupled prompt compared to PromptCast and Knowledge-Guided prompt across datasets and forecast horizons.}
\label{fig:forecasting_prompt}
\end{figure}

\begin{reviewtext}
We then performed an analysis to assess the robustness of our comparison. To obtain multiple predictions, we asked the model the same prompt for 50 replications, recorded the results at each replication, and derived the metrics for each of the sample predictions.  
Table~\ref{tab:pvalues_prompt_comparison} reports \(p\)-values from pairwise t-tests between the best prompt--covariate design and three alternatives (No Covariate, PromptCast, Knowledge-Guided) across datasets and forecast horizons for all evaluation metrics. We can see that the values are uniformly small on the test sets, indicating that the improvements are statistically reliable.  
For the Call Center dataset across both 1- and 7-step horizons, the \(p\)-values are all small, which shows that our method decisively outperforms all comparators. In the Influenza dataset, most test set \(p\)-values fall below \(10^{-4}\), with a few larger cases in the range \(10^{-2}\!-\!10^{-1}\). The validation \(p\)-values are sometimes larger (up to \(10^{-1}\)), but still provide clear evidence of improvement. For the Air Passengers dataset most results again show very small \(p\)-values, with only a few exceptions at short horizons where values remain small (on the order of \(10^{-3}\)). These observations confirm earlier findings in this section and in Section 6.2.
\end{reviewtext}

\begin{table}[H]
\centering
\caption{P-values from pairwise t-tests between the best prompt--covariate design and three alternatives--No Covariate, PromptCast, Knowledge-Guided--across datasets and forecast horizons.}
\label{tab:pvalues_prompt_comparison}
\resizebox{\textwidth}{!}{%
\begin{tabular}{l c l rrr rrr}
\toprule
\textbf{Dataset} & \textbf{Forecast Horizon} & \textbf{Prompt} &
\multicolumn{3}{c}{\textbf{Validation}} &
\multicolumn{3}{c}{\textbf{Test}} \\
\cmidrule(lr){4-6} \cmidrule(lr){7-9}
& & & \textbf{P-value (RMSE)} & \textbf{P-value (MAE)} & \textbf{P-value (MAPE)} & \textbf{P-value (RMSE)} & \textbf{P-value (MAE)} & \textbf{P-value (MAPE)}  \\
\midrule
\multirow{9}{*}{Influenza}
& \multirow{3}{*}{1}
  & No--Covariate           & \(\leq 10^{-4}\) & \(\leq 10^{-4}\) & \(\leq 10^{-4}\) & 2.03E-02 & \(\leq 10^{-4}\) & \(\leq 10^{-4}\) \\
& & PromptCast        & \(\leq 10^{-4}\) & \(\leq 10^{-4}\) & \(\leq 10^{-4}\) & \(\leq 10^{-4}\) & \(\leq 10^{-4}\) & \(\leq 10^{-4}\) \\
& & Knowledge-Guided  & \(\leq 10^{-4}\) & \(\leq 10^{-4}\) & \(\leq 10^{-4}\) & \(\leq 10^{-4}\) & \(\leq 10^{-4}\) & \(\leq 10^{-4}\) \\
\cmidrule(l){2-9}
& \multirow{3}{*}{2}
  & No--Covariate           & 1.06E-02 & \(\leq 10^{-4}\) & \(\leq 10^{-4}\) & 2.31E-02 & 1.09E-02 & 1.37E-01 \\
& & PromptCast        & 1.70E-02 & 1.50E-02 & 4.29E-02 & 1.11E-02 & 5.88E-03 & 8.08E-02 \\
& & Knowledge-Guided  & \(\leq 10^{-4}\) & \(\leq 10^{-4}\) & \(\leq 10^{-4}\) & \(\leq 10^{-4}\) & \(\leq 10^{-4}\) & \(\leq 10^{-4}\) \\
\cmidrule(l){2-9}
& \multirow{3}{*}{5}
  & No--Covariate           & 1.31E-01 & 1.15E-01 & \(\leq 10^{-4}\) & \(\leq 10^{-4}\) & \(\leq 10^{-4}\) & \(\leq 10^{-4}\) \\
& & PromptCast        & 3.39E-02 & 3.09E-02 & 8.52E-03 & 3.55E-01 & 2.99E-01 & 6.56E-02 \\
& & Knowledge-Guided  & \(\leq 10^{-4}\) & \(\leq 10^{-4}\) & \(\leq 10^{-4}\) & \(\leq 10^{-4}\) & \(\leq 10^{-4}\) & \(\leq 10^{-4}\) \\
\midrule
\multirow{6}{*}{Call Center}
& \multirow{3}{*}{1}
  & No--Covariate           & \(\leq 10^{-4}\) & \(\leq 10^{-4}\) & \(\leq 10^{-4}\) & \(\leq 10^{-4}\) & \(\leq 10^{-4}\) & \(\leq 10^{-4}\) \\
& & PromptCast        & \(\leq 10^{-4}\) & \(\leq 10^{-4}\) & \(\leq 10^{-4}\) & \(\leq 10^{-4}\) & \(\leq 10^{-4}\) & \(\leq 10^{-4}\) \\
& & Knowledge-Guided  & \(\leq 10^{-4}\) & \(\leq 10^{-4}\) & \(\leq 10^{-4}\) & \(\leq 10^{-4}\) & \(\leq 10^{-4}\) & \(\leq 10^{-4}\) \\
\cmidrule(l){2-9}
& \multirow{3}{*}{7}
  & No--Covariate          & \(\leq 10^{-4}\) & \(\leq 10^{-4}\) & \(\leq 10^{-4}\) & \(\leq 10^{-4}\) & \(\leq 10^{-4}\) & \(\leq 10^{-4}\) \\
& & PromptCast        & \(\leq 10^{-4}\) & \(\leq 10^{-4}\) & \(\leq 10^{-4}\) & \(\leq 10^{-4}\) & \(\leq 10^{-4}\) & \(\leq 10^{-4}\) \\
& & Knowledge-Guided  & \(\leq 10^{-4}\) & \(\leq 10^{-4}\) & \(\leq 10^{-4}\) & \(\leq 10^{-4}\) & \(\leq 10^{-4}\) & \(\leq 10^{-4}\) \\
\midrule
\multirow{9}{*}{Air Passengers}
& \multirow{3}{*}{1}
  & No--Covariate           & \(\leq 10^{-4}\) & \(\leq 10^{-4}\) & \(\leq 10^{-4}\) & \(\leq 10^{-4}\) & \(\leq 10^{-4}\) & \(\leq 10^{-4}\) \\
& & PromptCast        & 7.14E-03 & 2.23E-03 & 1.11E-02 & 2.88E-03 & 1.69E-03 & 6.74E-03 \\
& & Knowledge-Guided  & \(\leq 10^{-4}\) & \(\leq 10^{-4}\) & \(\leq 10^{-4}\) & 5.10E-04 & \(\leq 10^{-4}\) & \(\leq 10^{-4}\) \\
\cmidrule(l){2-9}
& \multirow{3}{*}{6}
  & No--Covariate          & \(\leq 10^{-4}\) & \(\leq 10^{-4}\) & \(\leq 10^{-4}\) & \(\leq 10^{-4}\) & \(\leq 10^{-4}\) & \(\leq 10^{-4}\) \\
& & PromptCast        & \(\leq 10^{-4}\) & \(\leq 10^{-4}\) & \(\leq 10^{-4}\) & \(\leq 10^{-4}\) & \(\leq 10^{-4}\) & \(\leq 10^{-4}\) \\
& & Knowledge-Guided  & \(\leq 10^{-4}\) & \(\leq 10^{-4}\) & \(\leq 10^{-4}\) & 7.05E-03 & 3.97E-05 & 9.82E-05 \\
\cmidrule(l){2-9}
& \multirow{3}{*}{12}
  & No--Covariate          & \(\leq 10^{-4}\) & \(\leq 10^{-4}\) & \(\leq 10^{-4}\) & \(\leq 10^{-4}\) & \(\leq 10^{-4}\) & \(\leq 10^{-4}\) \\
& & PromptCast        & \(\leq 10^{-4}\) & \(\leq 10^{-4}\) & \(\leq 10^{-4}\) & \(\leq 10^{-4}\) & \(\leq 10^{-4}\) & \(\leq 10^{-4}\) \\
& & Knowledge-Guided  & \(\leq 10^{-4}\) & \(\leq 10^{-4}\) & \(\leq 10^{-4}\) & \(\leq 10^{-4}\) & \(\leq 10^{-4}\) & \(\leq 10^{-4}\) \\
\bottomrule
\end{tabular}
}
\end{table}

\subsection{Sensitivity to Missing Covariates}
\begin{reviewtext}
Next, we examine how sensitive the forecasts are by analyzing the effect of random censoring of covariates on the evaluation metrics. Table~\ref{tab:all_censor_results} reports the forecast results when covariates are subjected to random censoring at different ratios. As the censoring level increases
from $0.1$ to $0.9$, errors generally increase across datasets and forecast horizons, although some fluctuations remain due to randomness in the censoring process. This
behavior reflects the loss of signal content in covariates as more entries are
masked.  

In the Influenza dataset, error growth is clear at higher forecast horizons. The validation and test RMSE more than double as the censoring level increases from $0.1$ to $0.9$. The performance of the Coupled prompt stays better than that of the No-Covariate prompt up to a censoring level of $0.9$ for forecast horizons of 1 and 2, but falls behind at the censoring level of $0.9$ for the forecast horizon of 5.  
In the Call Center dataset, degradation is also evident. At the forecast horizon of 1 with the censoring level of $0.9$, the No-Covariate prompt performs better, and a similar result appears at the censoring level of $0.5$ for the forecast horizon of 7.  
In the Air Passengers dataset, the error also increases as the level of censoring increases. After censoring levels of $0.5$, covariate integration no longer performs better than the No-Covariate prompt.  
In general, these results confirm that censored covariates reduce the accuracy of forecasts. The overall pattern shows the importance of careful covariate selection and reliable measurement to achieve stable and accurate LLM-based forecasts, while also showing that even partially observed covariates can still improve accuracy to some extent.

\end{reviewtext}
\begin{table}[H]
\centering
\caption{Forecasting results across datasets, forecast horizons, and censoring levels.}
\label{tab:all_censor_results}
\resizebox{\textwidth}{!}{%
\begin{tabular}{l c c rrr rrr}
\toprule
\textbf{Dataset} & \textbf{Forecast Horizon} & \textbf{Censoring Level} &
\multicolumn{3}{c}{\textbf{Validation}} &
\multicolumn{3}{c}{\textbf{Test}} \\
\cmidrule(lr){4-6} \cmidrule(lr){7-9}
& & & \textbf{RMSE} & \textbf{MAE} & \textbf{MAPE} (\%)& \textbf{RMSE} & \textbf{MAE} & \textbf{MAPE} (\%) \\
\midrule
\multirow{15}{*}{Influenza} 
  & \multirow{5}{*}{1}
    & 0.1 & 746.54  & 444.28 & 12.50 & 1367.23 & 557.80 & 17.25 \\
  & & 0.3 &  745.66  & 443.16 &  9.74 &  646.01 & 362.16 & 17.09\\
  & & 0.5 &944.42  & 536.64 & 10.29 & 3500.96 & 1196.60 & 48.25 \\
  & & 0.7 &1005.52 & 627.76 & 12.77 & 2441.39 & 869.64 & 18.74  \\
  & & 0.9 &   1153.40 & 661.96 & 12.15 & 1724.29 & 733.04 & 15.42 \\
\cmidrule(l){2-9}
  & \multirow{5}{*}{2}
    & 0.1 & 869.61  & 524.56 & 10.93 & 1890.55 & 829.00 & 17.93 \\
  & & 0.3 &1022.43 & 689.40 & 15.50 & 1876.15 & 836.52 & 28.11  \\
  & & 0.5 & 2569.13 & 1464.04 & 25.33 & 2899.08 & 1206.48 & 25.69 \\
  & & 0.7 & 2769.17 & 1701.28 & 35.63 & 2763.10 & 1281.64 & 30.39 \\
  & & 0.9 & 3072.41 & 1772.96 & 170.18 & 2333.31 & 1049.16 & 50.76 \\
\cmidrule(l){2-9}
  & \multirow{5}{*}{5}
    & 0.1 & 3125.41 & 2151.76 & 49.04 & 3265.41 & 1284.44 & 28.58 \\
  & & 0.3 &3458.45  & 2128.96 & 33.83 & 4681.76 & 1826.24 & 25.37  \\
  & & 0.5 &  5532.22  & 4045.76 & 442.86 & 2997.66 & 1379.00 & 34.11 \\
  & & 0.7 &  5461.74 & 4195.52 & 101.92 & 4769.56 & 2072.04 & 36.49\\
  & & 0.9 & 11420.23 & 7122.88 & 160.64 & 5007.93 & 1958.96 & 33.08 \\
\midrule

\multirow{10}{*}{Call Center} 
  & \multirow{5}{*}{1}
    & 0.1 &  75.29 &  53.48 & 11.61 & 142.09 &  85.86 & 21.10 \\
  & & 0.3 & 153.38 &  111.52 & 27.74 & 180.24 & 127.36 & 24.08 \\
  & & 0.5 & 144.32 &  93.62 & 24.79 & 172.35 & 124.29 & 34.55 \\
  & & 0.7 & 161.47 & 115.88 & 22.91 & 211.63 & 147.54 & 31.38 \\
  & & 0.9 & 185.16 & 133.72 & 25.44 & 227.92 & 169.37 & 34.61 \\
\cmidrule(l){2-9}
  & \multirow{5}{*}{7}
    & 0.1 &  79.38 &  54.33 & 11.87 & 112.45 &  76.79 & 17.01 \\
  & & 0.3 & 120.89 &  74.71 & 13.92 & 163.49 & 117.29 & 23.35 \\
  & & 0.5 & 173.12 & 135.57 & 35.05 & 229.78 & 185.64 & 48.91 \\
  & & 0.7 & 181.68 & 134.39 & 24.21 & 236.52 & 184.11 & 34.34 \\
  & & 0.9 & 203.75 & 153.84 & 28.62 & 269.47 & 204.22 & 38.19 \\
\midrule
\multirow{15}{*}{Air Passengers} 
  & \multirow{5}{*}{1}
    & 0.1 & 46.29 & 35.21 &  9.94 & 44.13 & 35.42 &  8.14  \\
  & & 0.3 & 46.69 & 37.33 & 10.53 & 55.62 & 43.92 & 10.02 \\
  & & 0.5 & 82.07 & 63.21 & 17.85 & 90.17 & 70.04 & 16.50 \\
  & & 0.7 &  70.61 & 54.96 & 14.98 & 60.84 & 44.50 &  9.76\\
  & & 0.9 & 79.48 & 63.42 & 17.60 & 87.10 & 72.04 & 16.59 \\
\cmidrule(l){2-9}
  & \multirow{5}{*}{6}
    & 0.1 & 43.29 & 37.71 & 10.37 & 49.34 & 41.29 &  8.86 \\
  & & 0.3 & 58.63 & 50.50 & 13.99 & 37.22 & 30.75 &  6.84 \\
  & & 0.5 &103.25 & 87.71 & 23.59 &117.42 & 91.54 & 21.74 \\
  & & 0.7 &  87.28 & 69.17 & 18.23 & 76.98 & 59.50 & 12.93\\
  & & 0.9 &100.26 & 86.00 & 23.80 &122.91 &107.29 & 25.56 \\
\cmidrule(l){2-9}
  & \multirow{5}{*}{12}
    & 0.1 & 25.21 & 21.00 &  5.67 & 51.78 & 34.75 &  8.55 \\
  & & 0.3 & 21.83 & 18.92 &  5.18 & 40.53 & 37.58 &  8.28 \\
  & & 0.5 &84.44 & 63.88 &  19.43 & 72.54 & 61.92 &  13.37 \\
  & & 0.7 & 102.17 & 84.21 & 24.53 & 86.35 & 76.17 & 17.53 \\
  & & 0.9 &124.60 &106.71 & 29.75 &119.74 & 88.42 & 20.99  \\
\bottomrule
\end{tabular}
}
\end{table}

\pagebreak

\subsection{Comparison with Existing Time Series Forecasting Methods}

\begin{reviewtext}
The previous sections focused on comparing different prompting strategies and 
covariate designs within the LLM-based framework. To place these results in 
context, it is also important to benchmark against established forecasting 
approaches. In particular, we compare our methods with ARIMA, a widely used 
classical statistical model, and LSTM, a standard deep learning baseline. 
This comparison allows us to assess whether covariate-informed LLM prompting 
can serve as a credible alternative to traditional forecasting methods.
Table~\ref{tab:combined_lstm_arima} reports the results for ARIMA and LSTM, while
Table~\ref{tab:all_datasets_prompt_comparison} summarizes the performance of our
prompting strategies. All experiments were carried out in the same setting,
using identical datasets, forecast horizons, and validation/test splits.
For the Influenza dataset, ARIMA performs poorly with very high errors
(e.g., 1-step test RMSE above 6600 and MAPE above 200\%), and LSTM achieves
better accuracy (1-step test RMSE 2766, MAE 1730). In contrast, the Coupled
prompt with covariates reduces the 1-step RMSE to 1343 and MAE to 540, cutting
error by more than 50\% compared to LSTM. At the longer horizon of 5 weeks,
Coupled prompt still achieves a test RMSE of 3805, far lower than ARIMA (7459) or LSTM
(6571), showing that LLM prompting remains competitive in noisy, high-variability
series.

For the Call Center dataset, LSTM achieves a test MAE of 49 in 1 step, and ARIMA
follows with 60. The Coupled prompt with day-of-week covariate improves this
to 56 MAE and 91 RMSE, which is better than ARIMA but slightly above LSTM for
short horizons. However, at the 7-step horizon, LSTM degrades to 82 RMSE and
ARIMA exceeds 100, while the Coupled prompt maintains only 87 RMSE and 52 MAE,
representing around 40\% lower error than ARIMA and stronger robustness than
LSTM in validation and test.
For the Air Passengers dataset, ARIMA is strong at long horizons: at 12 steps,
it reaches a test RMSE of 59 and MAPE of 11.7, while LSTM deteriorates to
RMSE 85 and MAPE 16.6. The Coupled prompt achieves RMSE 33 and MAE 30 at the
same horizon, cutting the error nearly by half relative to ARIMA and outperforming
LSTM by a large margin. At shorter horizons, such as 1 step, Coupled again
improves upon both classical models (test RMSE 43 vs. 78 for LSTM and 62 for
ARIMA). Overall, these results show that LLM-based prompting strategies are consistently
superior to ARIMA and competitive with LSTM across datasets.

\end{reviewtext}

\begin{table}[H]
\centering
\small
\setlength{\tabcolsep}{4pt}
\caption{Forecasting results of LSTM and ARIMA across datasets and forecast horizons.}
\label{tab:combined_lstm_arima}
\resizebox{\textwidth}{!}{%
\begin{tabular}{l c c ccc ccc}
\toprule
\textbf{Dataset} & \textbf{Step} & \textbf{Model} &
\multicolumn{3}{c}{\textbf{Validation}} &
\multicolumn{3}{c}{\textbf{Test}} \\
\cmidrule(lr){4-6} \cmidrule(lr){7-9}
& & & \textbf{RMSE} & \textbf{MAE} & \textbf{MAPE (\%)} & \textbf{RMSE} & \textbf{MAE} & \textbf{MAPE (\%)} \\
\midrule
\multirow{6}{*}{Influenza}
& \multirow{2}{*}{1} & LSTM  & 4848.98 & 3701.46 &  93.04 & 2766.92 & 1730.30 & 106.10 \\
&                     & ARIMA & 8593.71 & 7473.96 & 742.43 & 6603.52 & 4027.81 & 238.06 \\
\cmidrule(l){2-9}
& \multirow{2}{*}{2} & LSTM  & 5312.51 & 3845.87 &  78.40 & 3494.66 & 1423.81 &  56.22 \\
&                     & ARIMA & 7356.63 & 6313.16 & 632.88 & 6828.26 & 3691.23 & 185.82 \\
\cmidrule(l){2-9}
& \multirow{2}{*}{5} & LSTM  & 5959.22 & 4471.88 & 160.35 & 6571.31 & 3584.02 & 109.81 \\
&                     & ARIMA & 5624.28 & 4872.94 & 424.92 & 7459.18 & 5197.14 & 391.68 \\
\midrule
\multirow{4}{*}{Call Center}
& \multirow{2}{*}{1} & LSTM  &   66.46 &   50.53 & 11.08 &   78.85 &   49.17 & 12.05 \\
&                     & ARIMA &   72.69 &   58.89 & 14.04 &   82.27 &   60.20 & 14.68 \\
\cmidrule(l){2-9}
& \multirow{2}{*}{7} & LSTM  &   81.27 &   65.11 & 16.40 &   82.36 &   52.89 & 13.07 \\
&                     & ARIMA &   64.39 &   46.22 & 10.55 &  101.04 &   87.19 & 19.23 \\
\midrule
\multirow{6}{*}{Air Passengers}
& \multirow{2}{*}{1}  & LSTM  & 30.94 & 26.86 &  6.93 & 78.16 & 71.28 & 15.37 \\
&                      & ARIMA & 42.05 & 31.56 &  7.91 & 62.57 & 46.64 &  9.44 \\
\cmidrule(l){2-9}
& \multirow{2}{*}{6}  & LSTM  & 44.97 & 40.06 & 10.27 & 73.54 & 67.20 & 14.59 \\
&                      & ARIMA & 38.92 & 27.62 &  6.80 & 59.56 & 45.78 &  9.39 \\
\cmidrule(l){2-9}
& \multirow{2}{*}{12} & LSTM  & 27.62 & 21.28 &  5.31 & 84.71 & 77.09 & 16.58 \\
&                      & ARIMA & 14.65 &  9.75 &  2.45 & 59.63 & 53.03 & 11.68 \\
\bottomrule
\end{tabular}
}
\end{table}

\section{Conclusion}\label{sec:conclusion}

This study systematically analyzes the incorporation of covariates in time series forecasting with commonly used LLMs by proposing a set of candidate prompts and exploring key considerations and limitations. The proposed approach is implemented on three real-world datasets, and the results demonstrate that employing a well-structured prompting strategy, together with the inclusion of a suitable covariate from the dataset, can significantly enhance the accuracy and quality of forecasts.

Despite these promising results, several important challenges remain. First, many current LLM-based forecasting approaches lack formal mechanisms for uncertainty quantification and may struggle to generalize across diverse application domains. \textcolor{reviewblue}{A natural next step is to extend the framework to probabilistic forecasting, prompting for calibrated quantiles, prediction intervals, or full predictive distributions, and evaluating them with proper scoring rules and empirical coverage.} Moreover, most existing work is restricted to univariate time series, although multivariate forecasting has a wide applicability in real-world scenarios. \textcolor{reviewblue}{Generalizing the method to multivariate settings inspired by settings such as TimeCMA \citep{liu2025timecma}, conditioning on multiple synchronized series and cross-series covariates is a promising direction.}
Finally, evaluating LLM-based forecasts in real-time or streaming conditions would provide deeper insights into their robustness and operational viability. \textcolor{reviewblue}{Combining probabilistic outputs with multivariate conditioning in an online evaluation loop is a particularly promising direction for future work.}

\bibliographystyle{apalike}
\bibliography{ref}

\newpage
\appendix

\section*{Appendix}
In this part, we have included examples of our interaction with LLM.

\begin{figure}
    \centering
    \includegraphics[width=1\linewidth]{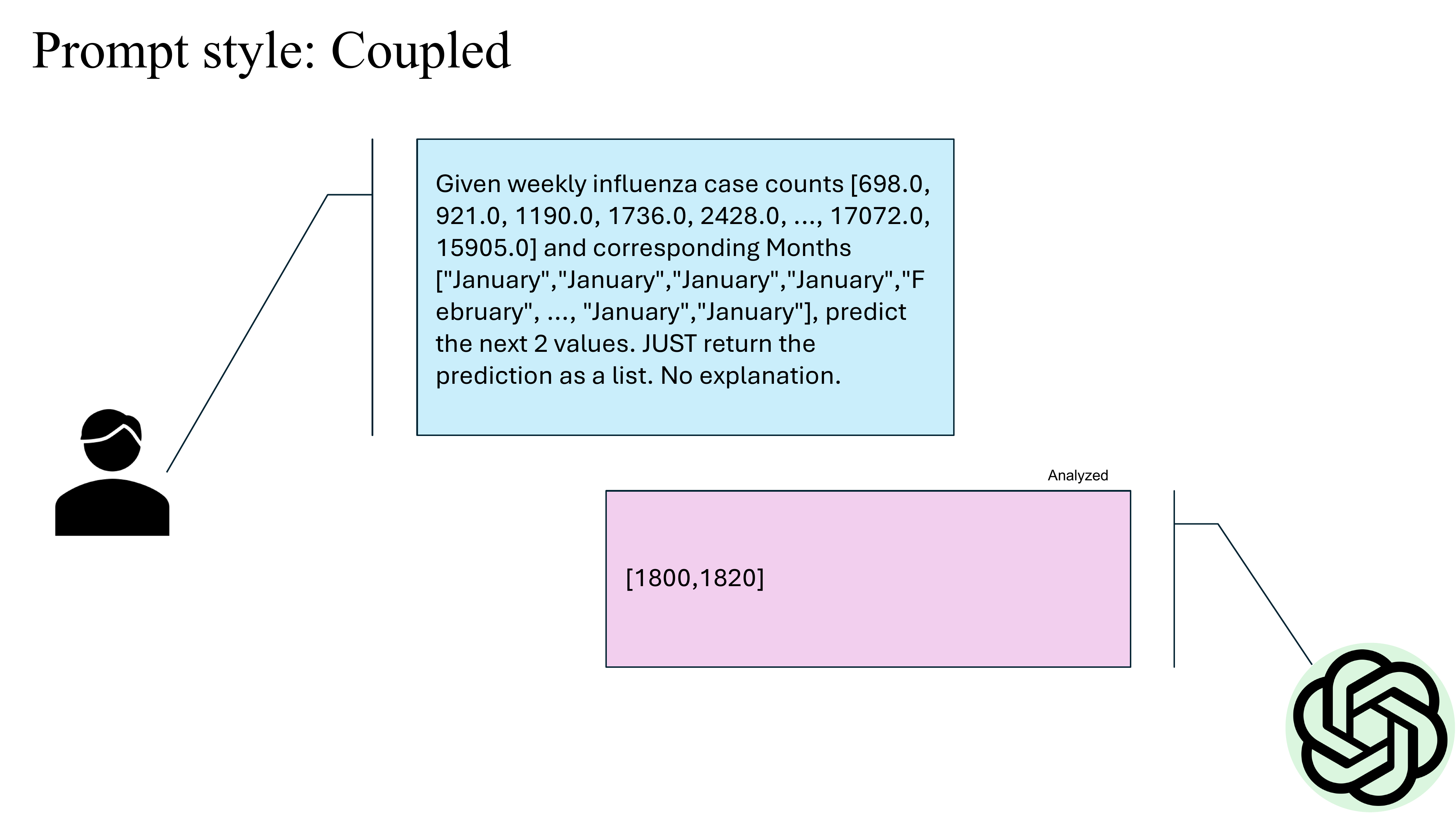}
    \caption{Illustration of the Coupled prompt style. Each observation is represented
as a key--value pair combining the target (weekly influenza case counts) with its
corresponding covariate (month). Future covariates are appended without targets,
and the LLM predicts the missing values in sequence.}
    \label{fig:prompt-coupled}
\end{figure}
\begin{figure}
    \centering
    \includegraphics[width=1\linewidth]{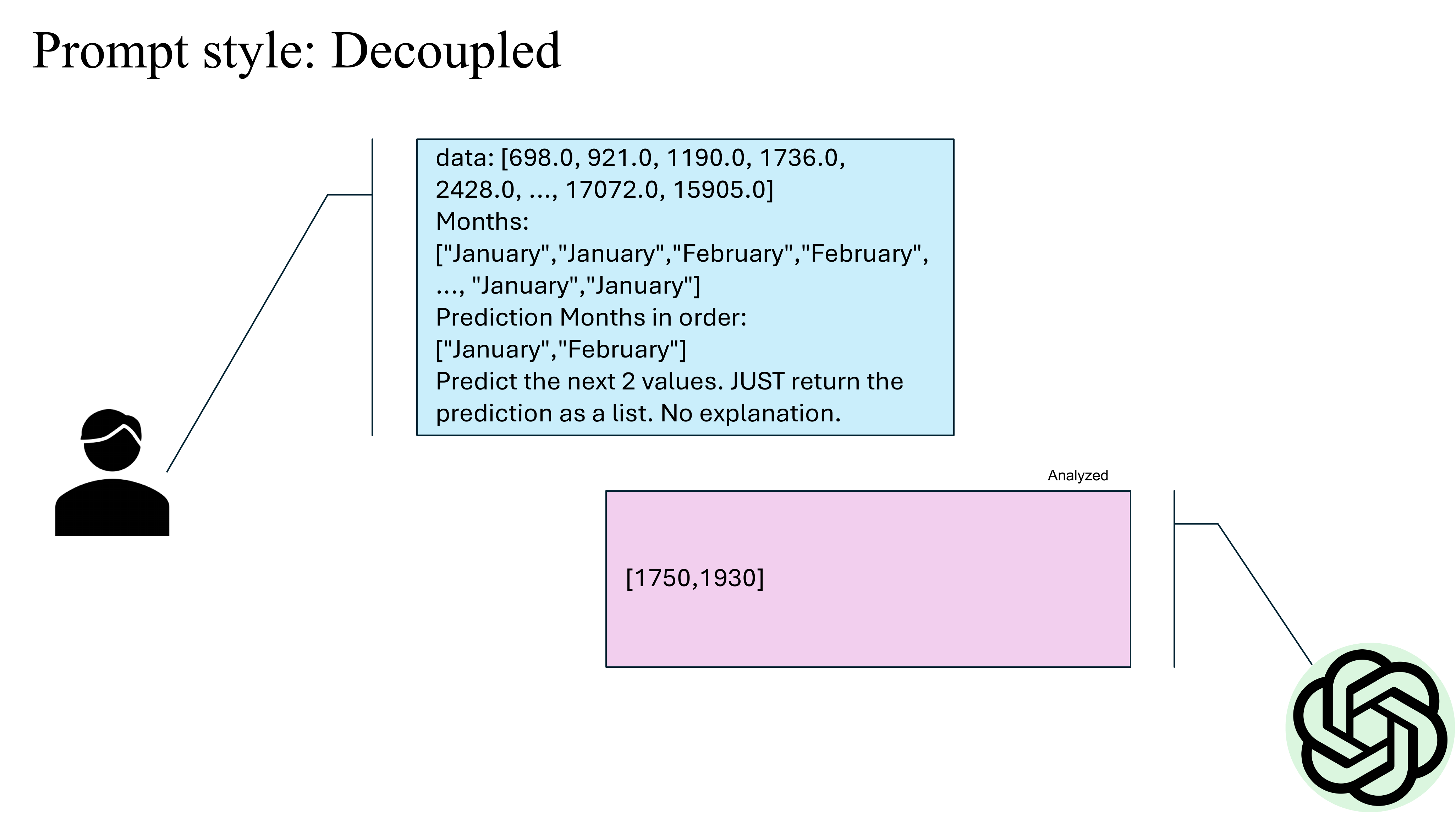}
    \caption{Illustration of the Decoupled prompt style. Target values (weekly influenza
case counts) and covariates (months) are presented in separate lists, with future
covariates provided explicitly. The LLM uses these aligned sequences to generate
the next predictions.}

    \label{fig:placeholder}
\end{figure}
\begin{figure}
    \centering
    \includegraphics[width=1\linewidth]{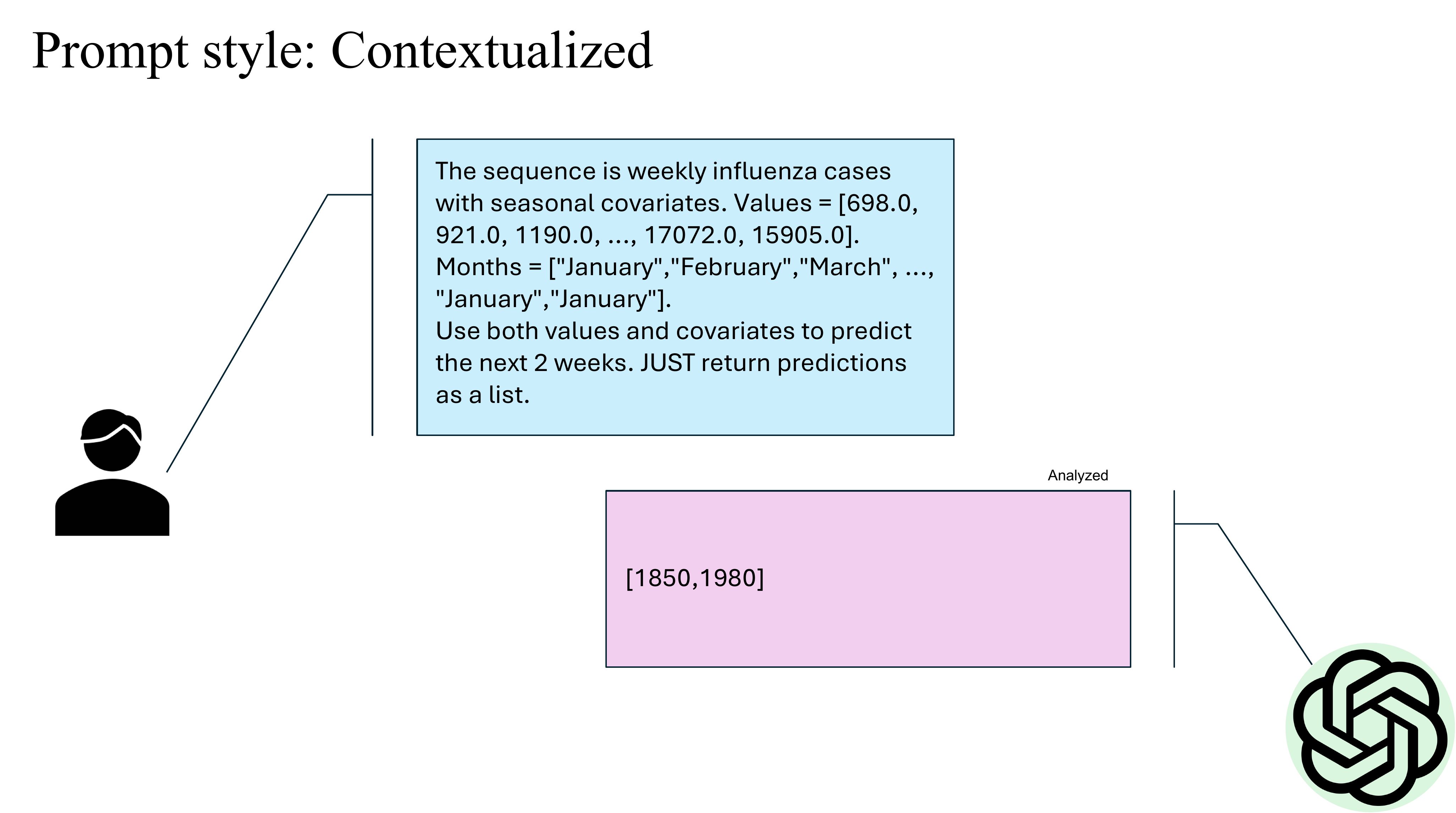}
    \caption{Caption}
    \label{fig:placeholder}
\end{figure}
\begin{figure}
    \centering
    \includegraphics[width=1\linewidth]{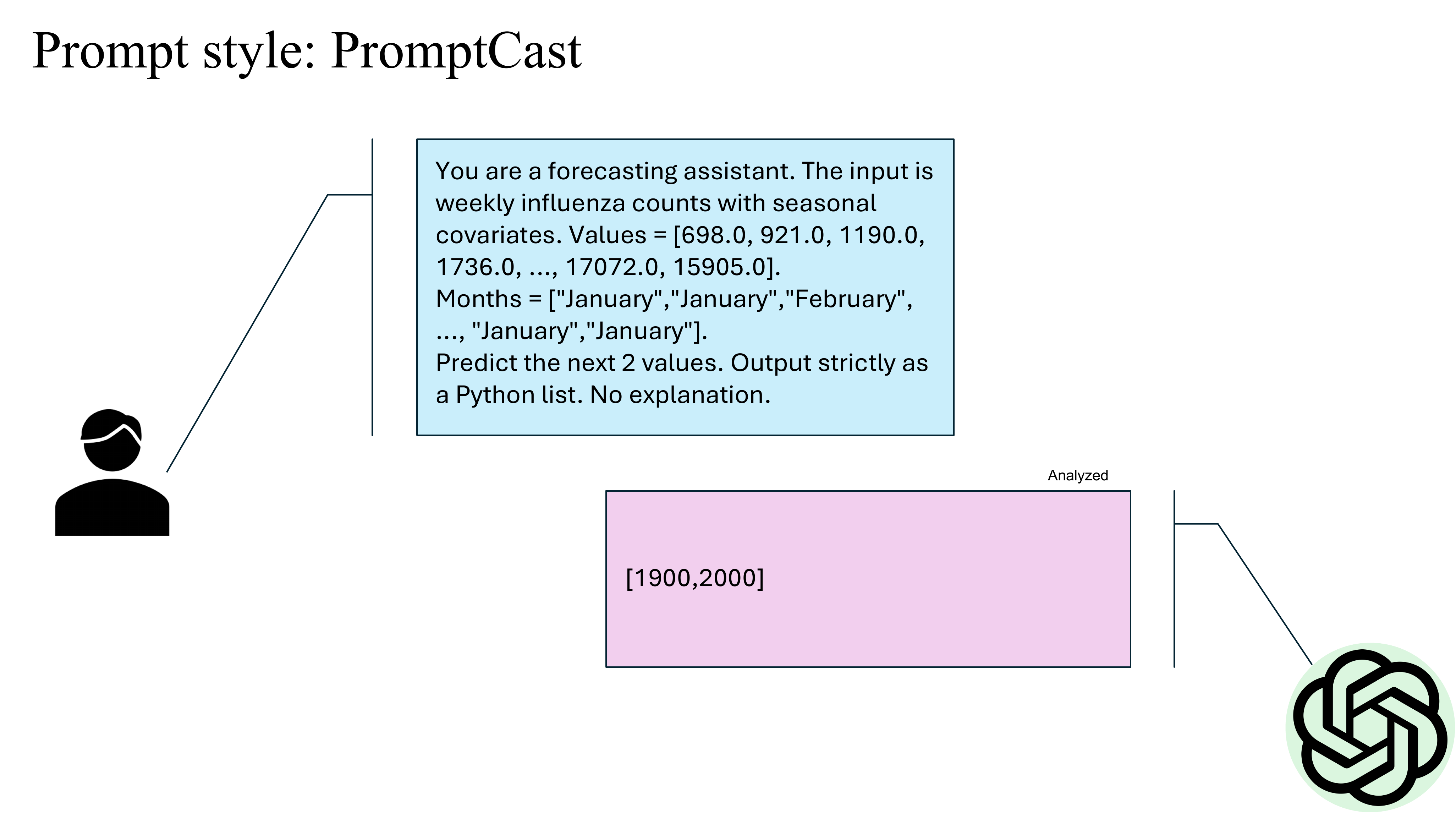}
    \caption{Caption}
    \label{fig:placeholder}
\end{figure}
\begin{figure}
    \centering
    \includegraphics[width=1\linewidth]{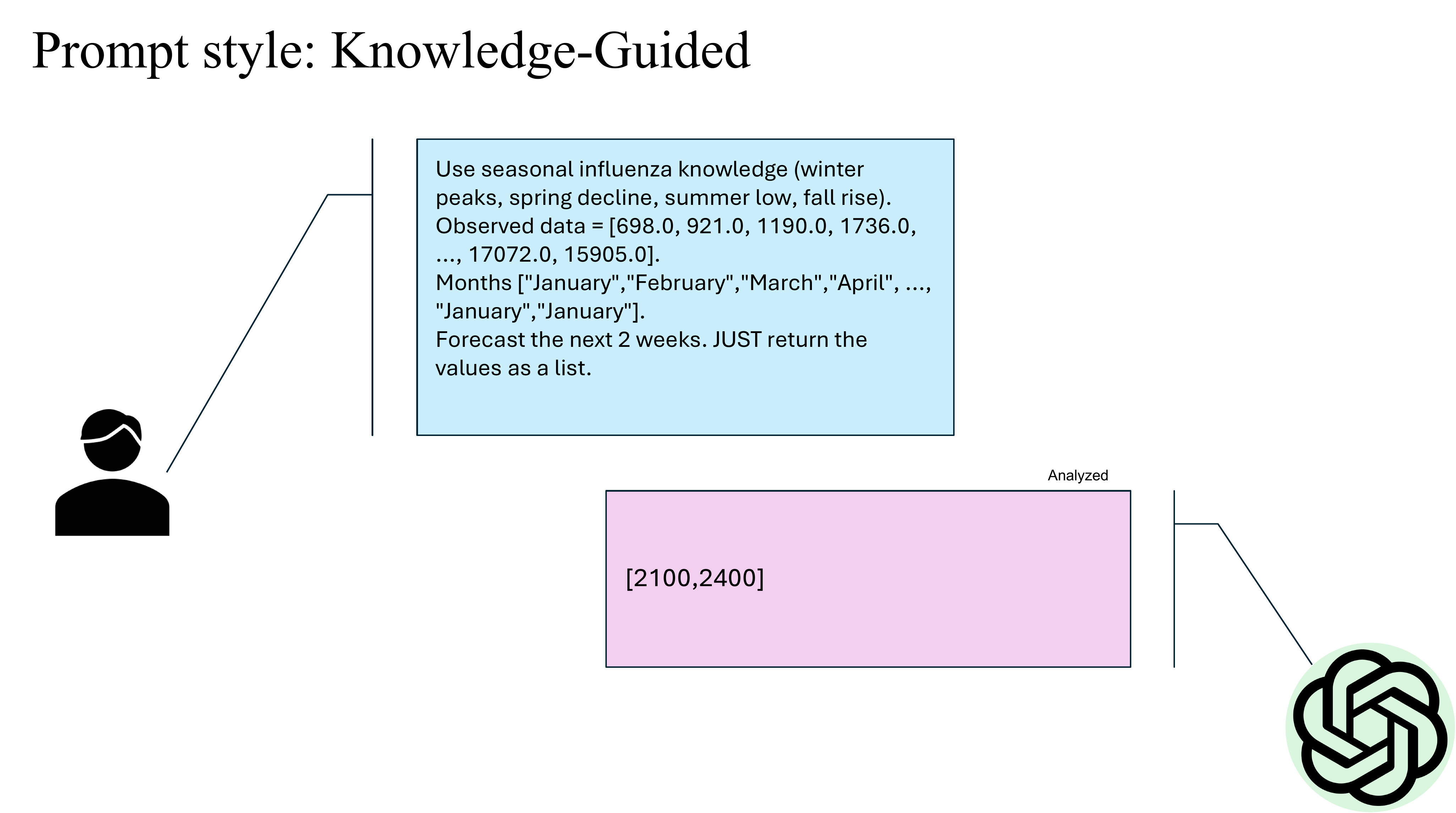}
    \caption{Caption}
    \label{fig:placeholder}
\end{figure}

\pagebreak

\begin{figure}[H]
    \centering
    \includegraphics[width=1\linewidth]{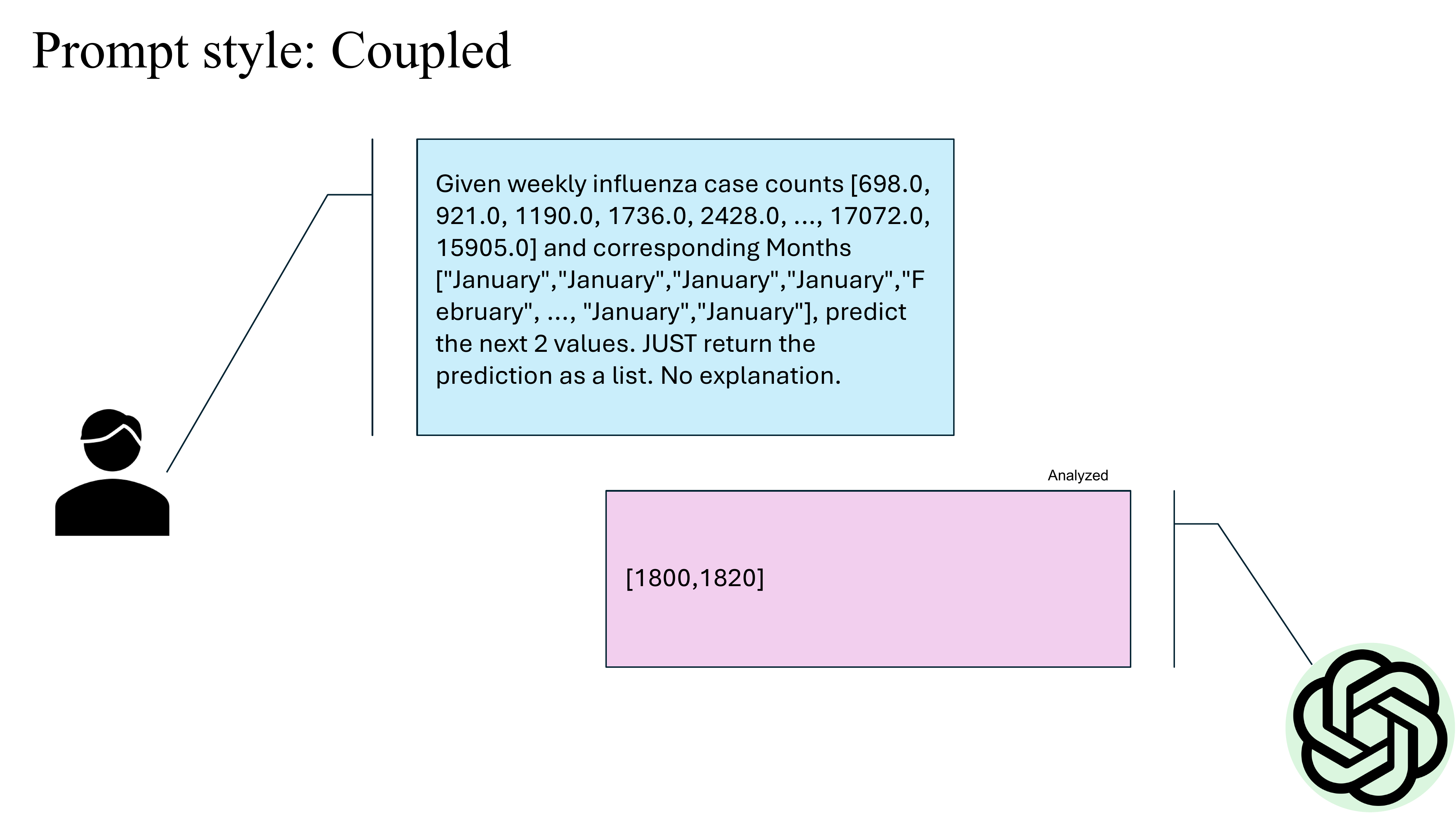}
    \caption{Illustration of the Coupled prompt style. Each observation is represented as a key--value pair combining the target (weekly influenza case counts) with its corresponding covariate (month). Future covariates are appended without targets, and the LLM predicts the missing values in sequence.}
    \label{fig:prompt-coupled}
\end{figure}

\begin{figure}
    \centering
    \includegraphics[width=1\linewidth]{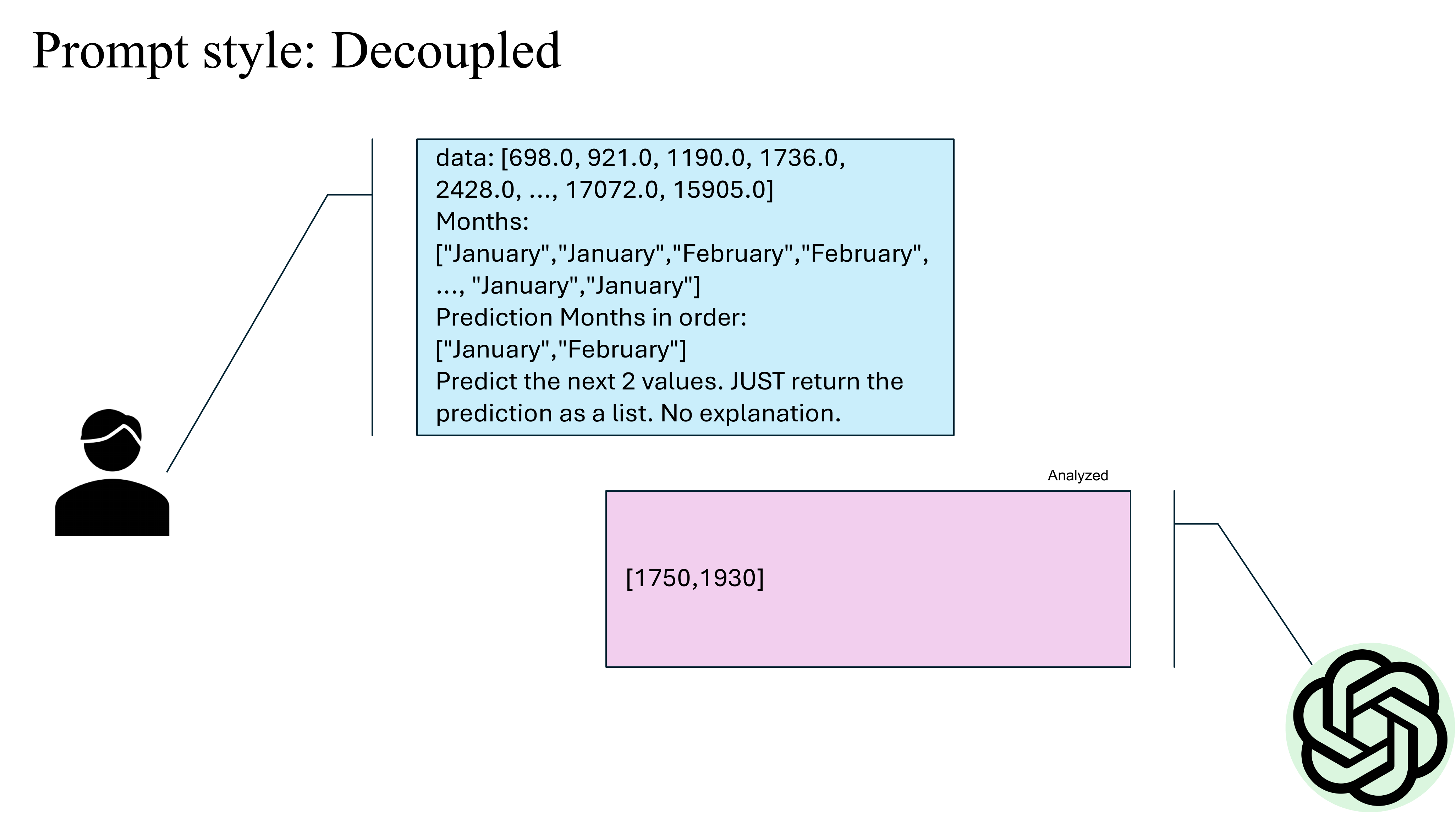}
    \caption{Illustration of the Decoupled prompt style. Target values and covariates are presented in separate lists, with future covariates provided explicitly. The LLM uses these aligned sequences to generate the next predictions.}
    \label{fig:prompt-decoupled}
\end{figure}

\begin{figure}
    \centering
    \includegraphics[width=1\linewidth]{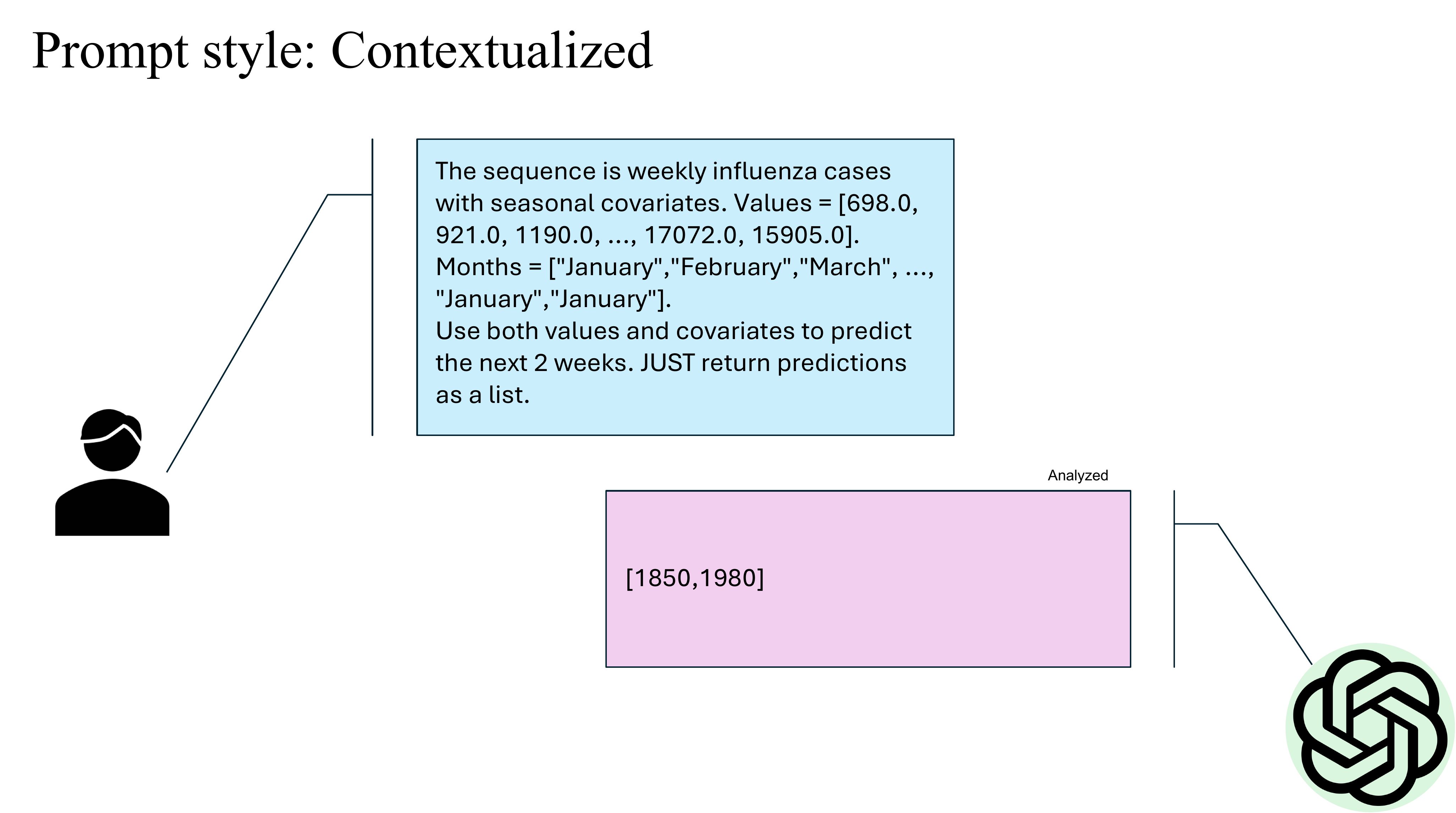}
    \caption{Illustration of the Contextualized prompt style. Historical values and covariates are supplemented with descriptive context (e.g., seasonality cues). The LLM leverages both numerical inputs and semantic hints to generate future predictions.}
    \label{fig:prompt-contextualized}
\end{figure}

\begin{figure}
    \centering
    \includegraphics[width=1\linewidth]{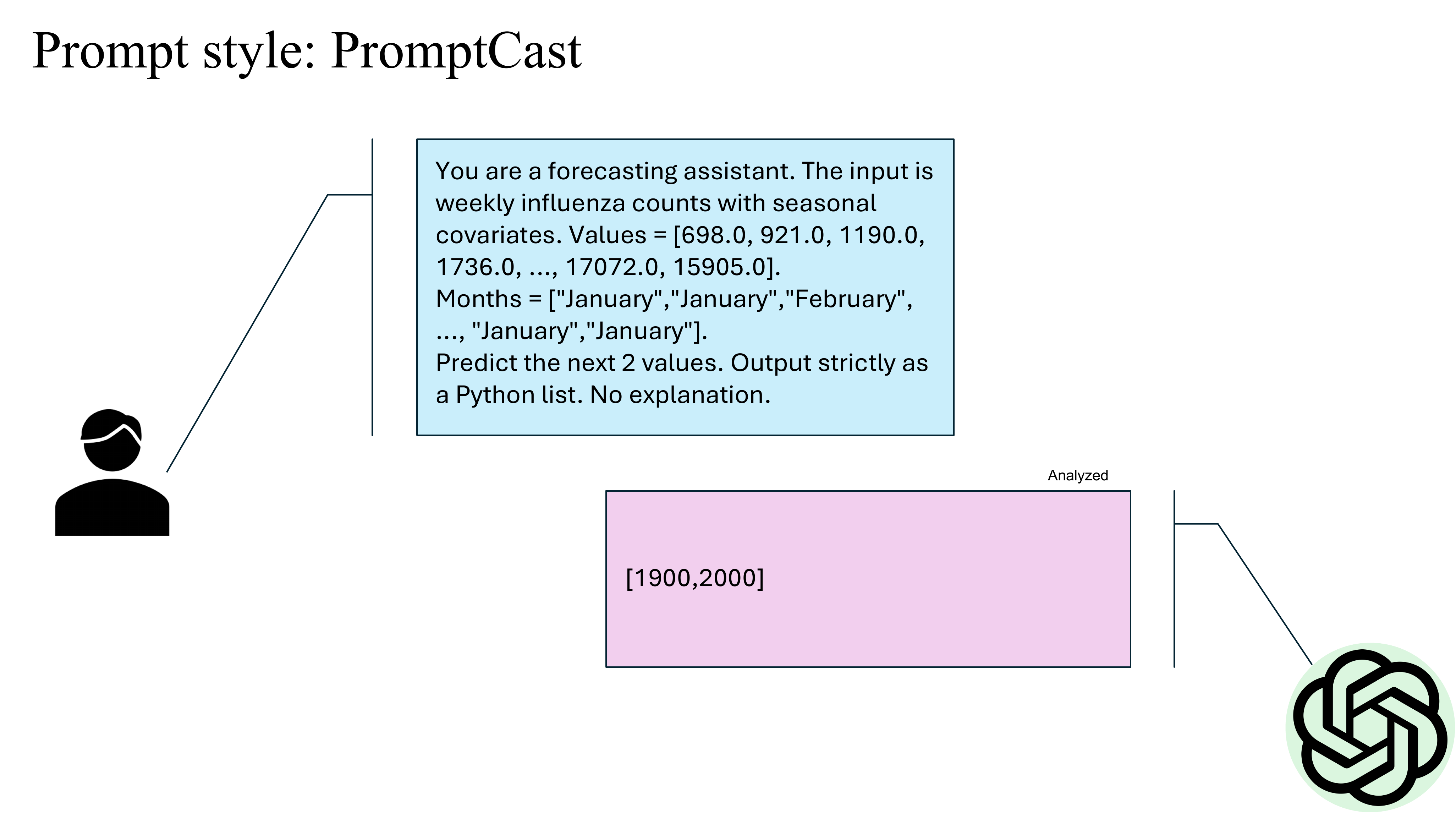}
    \caption{Illustration of the PromptCast baseline style. The LLM is instructed as a forecasting assistant, with inputs provided as values and covariates, and outputs constrained to a Python-style list without explanation.}
    \label{fig:prompt-promptcast}
\end{figure}

\begin{figure}
    \centering
    \includegraphics[width=1\linewidth]{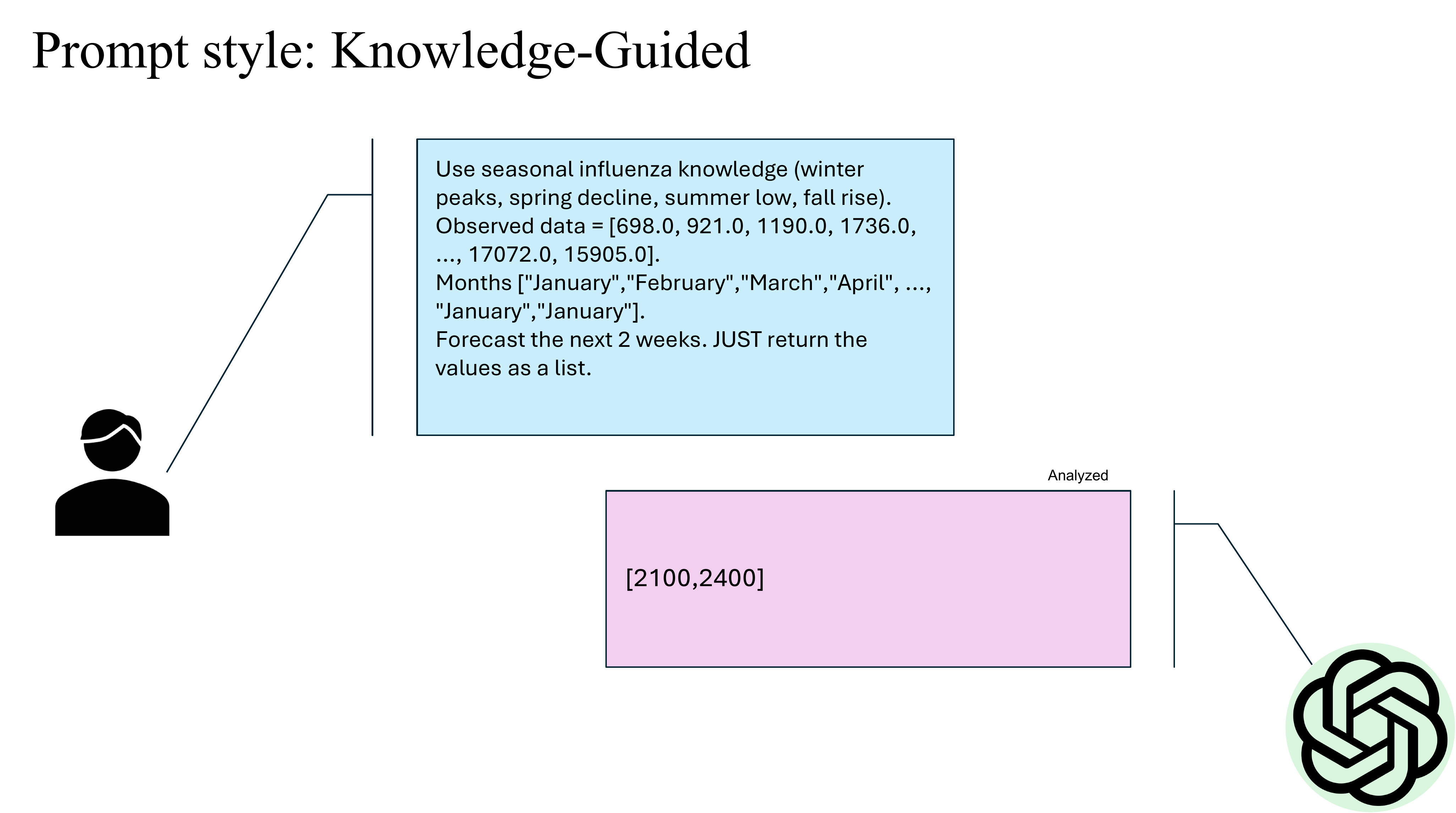}
    \caption{Illustration of the Knowledge-Guided prompt style. Domain knowledge about seasonal patterns (e.g., winter peaks, summer lows) is included alongside observed values and covariates, guiding the LLM to incorporate prior knowledge into its forecasts.}
    \label{fig:prompt-knowledge}
\end{figure}
\end{document}